\documentclass[twoside,11pt]{article}

\usepackage{blindtext}

\usepackage[abbrvbib, preprint]{jmlr2e} 
\graphicspath{{figures}} 
\usepackage{amsmath,amssymb}
\usepackage[table]{xcolor}
\usepackage{longtable}
\usepackage{hhline}
\usepackage{hyperref}
\hypersetup{ hidelinks }
\newtheorem{thm}{Theorem}[section]
\newtheorem{defn}[thm]{Definition}

\newcommand{\inv}{^{-1}}

\newcommand{\R}{\mathbb{R}}

\usepackage{lastpage}
\jmlrheading{25}{2024}{1-\pageref{LastPage}}{8/24}{PUBLISHED-DATE}{PAPER-ID}{Amish Mishra and Francis Motta} 

\ShortHeadings{Topological Feature Learning for Protein Stability}{Mishra and Motta}
\firstpageno{1}

\begin{document}

\title{A Pipeline for Data-Driven Learning of Topological Features with Applications to Protein Stability Prediction}

\author{\name Amish Mishra \email amish\_mishra@taylor.edu \\
       \addr Mathematics Department\\
      Taylor University\\
       Upland, IN 46989, USA
       \AND
       \name Francis Motta \email fmotta@fau.edu \\
       \addr Department of Mathematical Sciences\\
       Florida Atlantic University\\
       Boca Raton, FL 33431, USA}

\editor{My editor}

\maketitle

\begin{abstract}
    In this paper, we propose a data-driven method to learn interpretable topological features of biomolecular data and demonstrate the efficacy of parsimonious models trained on topological features in predicting the stability of synthetic mini proteins. We compare models that leverage automatically-learned structural features against models trained on a large set of biophysical features determined by subject-matter experts (SME). Our models, based only on topological features of the protein structures, achieved 92\%-99\% of the performance of SME-based models in terms of the average precision score. By interrogating model performance and feature importance metrics we extract numerous insights that uncover high correlations between topological features and SME features.
    We further showcase how combining topological features and SME features can lead to improved model performance over either feature set used in isolation, suggesting that, in some settings, topological features may provide new discriminating information not captured in existing SME features that are useful for protein stability prediction. 
\end{abstract}


\section{Introduction}

As proteins fold from an unfolded state into their tertiary conformation(s), they lose free energy through various mechanisms (e.g, van der Waals and electrostatic interactions, hydrophobic collapse, etc.) \citep{Stollar2020-mw} and the multitude of weak interatomic interactions between their constitute atoms \citep{rocklin}. The relative quantity of lost free energy may be regarded as a measure of a conformations stability \citep{rapid_protein_stab_pred}. 
Whether they are natural or synthetic (designed/engineered), many proteins require stable structures to maintain a functional tertiary structure. For instance, some diseases in humans can be explained by the moderately weak stability of the biologically function state that may lead to misfolding or destabilization \citep{NIELSEN2020111}. Thus, understanding the determinants of stability is an important problem in structural biology with implications to human health and the design of functionally stable proteins \citep{rocklin}.

Recently, machine learning (ML) has been utilized as a promising tool in the prediction of protein stability. Approaches have included the use of the primary sequence and the biophysical interactions between the atoms \citep{rapid_protein_stab_pred, protein_stab_prediction_mutation, predicting_changes_in_protein_thermodynamic_stability, elnaggar2021prottrans}. Some work has also been done on topological feature engineering from protein structure. In \citep{CangMuWuOpronXiaWei_2015}, the authors develop a molecular topological fingerprint-based support vector machine (MTF-SVM) classifier that leverages persistent homology to achieve high accuracy in protein classification tasks. In \citep{Jiang2023.08.25.554762}, the authors combine topological data analysis (TDA) and geometric deep learning (GDL) to analyze protein binding pockets, providing a comprehensive understanding of protein structural motifs by integrating both local and global representations. In \citep{swenson2020persgnn}, PersGNN is introduced, a hybrid deep learning model combining graph neural networks and topological data analysis to improve protein function prediction from structural data.


Previously, the work done in \citep{rocklin} exposed biophysical and energetic determinants of stability. In this paper we take a data-driven approach to search for additional interpretable, topological features associated with stability.  


Our approach involves calculating topological invariants in the atomic arrangements of protein tertiary structures, which we represent as persistence diagrams (PDs), and then identifying and interpreting the most informative regions of the PDs.  By building simple machine learning models trained on topological features and/or on biophysical features, which are known by subject-matter experts (SME) to be associated with protein (in)stability, we compared the capacity of learned topological features to predict stability in the hopes of contributing new structural determinants of stability. These analyses led to the following observations:

\begin{itemize}
    \item Topological features-based ML models achieved 92\%-99\% of the performance of models trained on features identified by subject-matter experts in modeling the stability of proteins with secondary structures $\alpha \alpha \alpha$, $\beta \alpha \beta \beta$, $\alpha \beta \beta \alpha$, and $\beta \beta \alpha \beta \beta$. (Analyzed from Table~\ref{tab:cder_sme_rfc}.)
    \item The incorporation of topological features in model development demonstrated a statistically significant improvement in the classification of $\alpha \beta \beta \alpha$ and $\beta \beta \alpha \beta \beta$ protein design topologies with regard to the average precision score (APS) metric for classification. (See Table~\ref{tab:cder_sme_rfc}.)
    \item Strong correlations ($|r|>0.9$) were identified between SME features known to be determinants of stability and the learned topological features. In fact, we observed a general trend that the topological features that were important for classifying stable/unstable protein designs were highly correlated with SME features that were independently important for classifying stable/unstable protein designs. (Illustrated in Figures~\ref{fig:HHH_corr} to~\ref{fig:EEHEE_corr}.)
\end{itemize}

To learn new, informative topological descriptors, we adopted a data-driven approach designed to identify the topological differences between the most and least stable designed mini-proteins reported in \citep{rocklin}. Our approach made use of an adaptive template system described in detail in Section~\ref{sec:prelim}: Cover-Tree Differencing via Entropy Reduction (CDER) \citep{cover_trees} to identify topological structures that were overrepresented either in stable or unstable designs. These small numbers of regions were then used to vectorize PDs into finite-dimensional feature vectors that were used to train stability prediction models. 

The remainder of the paper is divided as follows. Section~\ref{sec:data_context} provides context of the data and how it was preprocessed before doing our experiments. Section~\ref{sec:cder_application} explains how CDER was used for feature engineering, how models were trained, and what analyses and conclusions we could draw from the tables and figures generated. Finally, Section~\ref{sec:discussion} summarizes this work and our results. 

\section{Preliminaries}\label{sec:prelim}

Topological Data Analysis (TDA) techniques have gained prominence in recent years for their ability to extract and analyze complex geometric and topological structures within data sets. The concept of Persistent Homology (PH) has proven to be a fruitful tool in extracting latent features from data sets by summarizing topological structures across scale in families of simplicial complexes and representing them compactly as collections of real-number pairs (\textit{persistence pairs}) in a so-called persistence diagram (PD). The PD is a multiset of points in $\R^2$, however this representation is inconvenient when attempting to use PDs in machine learning (ML) pipelines. Thus, developing methods of vectorizing the PD has become an active area of research \citep{pers_land, Reininghaus,Bendich,adams2017persistence, Chung_frontiers_hr_ml, poly_pers_diagram}. Most of these methods take the form of a (parameterized) transformation, $T(D; \Theta)$ from the space of persistence diagrams to $\mathbb{R}^n$. The choice of vectorization, and of the (hyper)parameters in each methods allows flexibility in the ultimate topological feature engineering. However, the best choice of method and parameters is not always clear, although machine learning techniques such as hyperparameter optimization can be used to learn good choices of PD vectorization parameters for a particular modeling task \citep{mottahyperparam2019}. For example, the persistence images vectorization \citep{adams2017persistence} allow a modeler to choose a region of the PD plane to focus on, and the dimension of the vectorization, however the region is constrained to be rectangular, which may result in low-information components and highly correlated features. 

In this manuscript, we take a different data-driven approach to featurizing the persistence diagrams for a supervised machine learning task. Instead of using model performance metrics to guide the choice of hyperparameters of the vectorization, we seek first to identify the most discriminating regions of the diagrams themselves; those portions of the PD that are discriminating between the classes of interest in our dataset. Thus, we let the data decide for us which portions of the PD to use for the machine learning task. The regions which are most discriminating are automatically learned by a measure of local entropy in the labels of persistence pairs, assigned according to the label of the data sample which generated the associated persistence diagram. 

\subsection{Persistent Homology}

In algebraic topology, homology serves as an invariant that captures information about the ``holes" present within a space. For a given topological space $X$, the concept of the $k$-dimensional holes (such as connected components, loops, and enclosed volumes) is mathematically represented by the $k$-th homology group, denoted as $H_k(X)$. The number of independent $k$-dimensional holes in $X$ is quantified by the rank of this group, referred to as the $k$-th Betti number, $\beta_k$. A thorough exploration of homology can be found in \citep{hatcher2002algebraic}. For finer details of the topics of this section, see Section~\ref{sec:extra_persistence}.

Considering a sequence of nested topological spaces $\{X_1 \subseteq X_2 \subseteq \cdots \subseteq X_n\}$, the inclusion $X_i \subseteq X_{j}$ for $i \leq j$ induces a linear map $H_k(X_i) \rightarrow H_k(X_{j})$ on the $k$-th homology groups for all $k \geq 0$. Persistent homology is a framework that monitors the evolution of elements in $H_k(X_i)$ as the index $i$, which can be interpreted as a scale parameter, increases. This concept has been extensively studied in recent works \citep{Edelsbrunner, Edelsbrunner2000TopologicalPA, Zomorodian2005, Zomorodian_comp_top, 2009carlsson}.

Often, and for all data considered in this study, a dataset is a finite point cloud in some metric space. To construct a nested sequence of topological spaces from such a dataset, numerous methods have been proposed including the Vietoris-Rips filtration \citep{Edelsbrunner}, the (weighted) Alpha filtration  \citep{Edelsbrunner}, and the Delaunay-Rips filtration \citep{mishra_motta_frontiers}. We utilize the weighted alpha implemented in the Geometry Understanding in Higher Dimensions (GUDHI) library \citep{gudhi}, which allow us to incorporate into the evolution of topological structures both the locations of atoms in protein conformations and atomic identities. 

As stated, a standard representation of persistent homology is through a PD, which is a multiset of points in the plane $\mathbb{R}^2$. For a fixed homological dimension $k$, each topological feature is associated with a point $(x, y)$ (a persistence pair) where $x$ and $y$ denote the indices at which the feature first appears (births) and subsequently disappears (dies), respectively. Since no topological feature can disappear before it appears, all points in a PD lie on or above the diagonal $y = x$. In a PD, distinct topological features might share the same birth and death coordinates, leading to multiple points at the same location. Features close to the diagonal are often regarded as noise, while those further away are considered more significant topological attributes. 

\subsection{PH and CDER}

\begin{figure}[ht]
    \centering
    \includegraphics[width=\columnwidth]{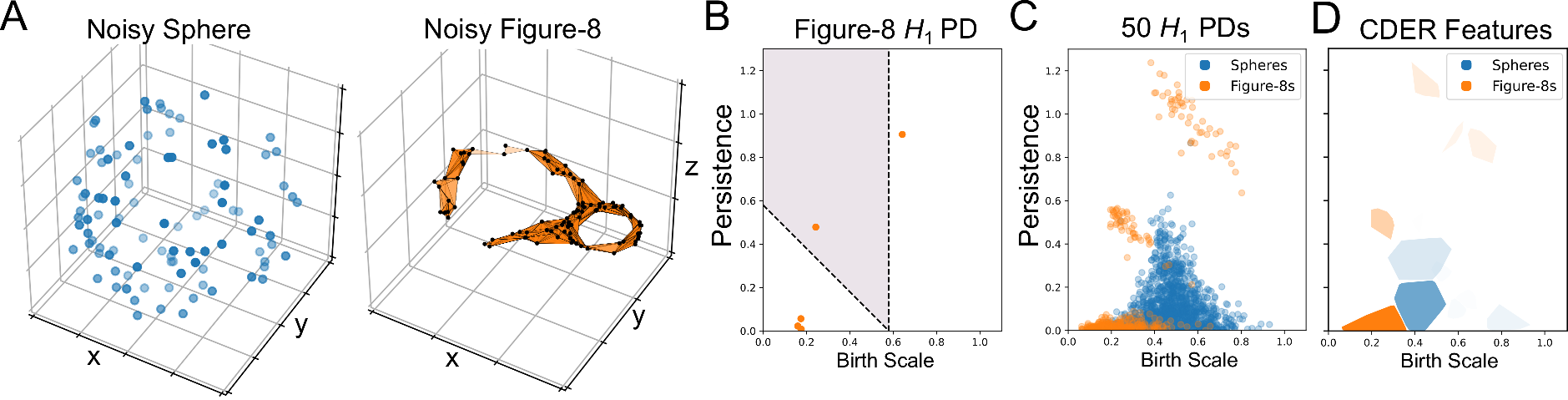}
    \caption{Toy Example of TDA-CDER pipeline. (A) Example of noisy point clouds sampled from a sphere (left) and a figure-8 (right). Orange triangles show structure added to the figure-8 by connecting points that are within a certain scale (0.58). (B) $H_1$ PD of the noisy figure-8 in (A), which shows the presence of topological holes and their persistence over scales. The point in the shaded region corresponds to the smaller hole already formed at the scale  (0.58) indicated in (A). The higher persistence point to the right of the shaded region will be ``born'' as the scale parameter increases and closes the open loop in the noisy figure-8. At a large enough scale, the small hole will ``die'' as the hole is filled in. The persistence of a topological feature is the difference in its birth and death scales. (C) $H_1$ PDs of 50 of each randomly sampled noisy spheres and figure-8s. (D) Features of the PDs learned by CDER to be discriminating between spheres and figure-8s. CDER ignores regions with points common in spheres and figure-8s. }
    \label{fig:[h_cder_pipeline]}
\end{figure}

To illustrate the topological vectorization approach implemented in this study, we present  in Figure~\ref{fig:[h_cder_pipeline]} a cartoon example of how CDER can be used to learn discriminating regions of PDs, and generate parsimonious topological feature vectors to be used in downstream ML model training. Consider a binary classification task of discriminating between two shape classes: spheres and figure-8s. Each dataset sample is a labeled point cloud in $\mathbb{R}^3$ gotten by drawing random points from one of the two underlying shapes (with noise). Figure~\ref{fig:[h_cder_pipeline]}A shows one dataset sample labeled as a sphere (blue points) and one labeled as a figure-8 (orange points). Each such data sample is transformed into a nested family of simplicial complexes by connecting points within a distance $s \in [0,\infty)$. As the scale parameter $s$ is allowed to grow, more points are connected, and the topological structures are tracked in the PD. Figure~\ref{fig:[h_cder_pipeline]}B shows the $H_1$ PD, which tracks 1-dimensional loops, for the example noisy figure-8 sample. The shaded region corresponds to the simplicial complex overlaid on the point cloud at the scale indicated in Figure~\ref{fig:[h_cder_pipeline]}A. By collecting all such labeled diagrams for all labeled samples, we observe discernible concentrations of orange and blue persistence pairs, Figure~\ref{fig:[h_cder_pipeline]}C. This suggests the existence of an (unknown) distribution corresponding to persistence pairs that encode topological features of noisy samples of the figure-8 space, and separately a distribution corresponding to the sphere. Importantly, much of the PD appears to not support any persistence pairs for these shape classes, some regions are distinctly one or the other, while other regions contain pairs from both classes. By applying CDER to the labeled point clouds in Figure~\ref{fig:[h_cder_pipeline]}C, a small set of discriminating regions are automatically identified Figure~\ref{fig:[h_cder_pipeline]}D. Each of these will correspond to a topological feature in the resulting vectorization. Whenever another (unlabeled) persistence diagram is generated, it can be evaluated in relation to these learned CDER features. 

\subsection{Cover-Tree Differencing via Entropy Reduction (CDER)} \label{sec:cder_theory}
Template systems are typically constructed by defining a set of template functions $\{f_i\}_{i=1}^m$, where each function $f_i$ is a continuous real-valued function defined on the space of interest, which in our case is a region in $\mathbb{R}^2$ containing the persistence pairs of PDs. Template functions are chosen to capture the local structure of the data set, and are typically constructed using a fixed set of parameters that are chosen based on prior knowledge or heuristics \citep{perea2023template}. 

In contrast, adaptive template systems aim to construct the template functions in a way that is tailored to the specific problem at hand. This is typically done by incorporating prior knowledge about the data set into the construction of the template functions, such as the distribution of the data points or the geometry of the underlying space \citep{adaptive_template}. 

Once template functions are constructed on PDs, they can be used to vectorize the diagram to be used as input for an ML algorithm. Typically, we think of each template function $\{f_i\}_{i=1}^m$ as giving rise to a coordinate in a vectorization.  

We limit our discussion in this section to the context of supervised learning tasks. Let $\mathcal{X} = \{X_1, X_2,\dots, X_N\}$ be a collection of point clouds where each $X_i \subset \R^D$ for fixed $D$ and the number of points in each point cloud can vary. Let $\lambda: \mathcal{X} \to \Lambda$ be a function that assigns a label in $\Lambda = \{1,\dots,L\}$ to each point cloud in $\mathcal{X}$. The goal is to identify compact and convex regions $\Omega \subset \R^D$ that have a higher density of points from one label class in $\Lambda$ than any other. For example, if a region $\Omega$ has a much higher density of points labeled as ``2'' than any other label, $\Omega$ would be associated with the label ``2''. Now, given a region $\Omega \subset \R^D$, what measure can be used to identify if the region is characteristically dense in a certain label or not?

The notion of entropy in the information theoretic sense is helpful in this endeavor. The regions $\Omega$ where entropy is high can be treated as non-informative regions; where entropy is low, identify that region with the label of the dominating point cloud.

To make this more formal, begin by defining
\[\underline{\mathcal{X}} := \bigcup_{i=1}^N X_i\]
as the set of all points from all point clouds. There is a natural map, $i: \underline{\mathcal{X}} \to \mathcal{X}$, to identify the points of $\underline{\mathcal{X}}$,
\[i(x) := X_{i},\ \text{if } x \in X_i.\]
Now define the set of all points in $\underline{\mathcal{X}}$ labeled as $l \in \Lambda$ by
\[\underline{\mathcal{X}}_l := i\inv \circ \lambda\inv (l).\]
For each $x \in \underline{\mathcal{X}}$, assign weights $w(x)$ by the following procedure.
\begin{enumerate}
    \item When attempting to make adaptive template functions for a test set, assume the labels from $\Lambda$ have equal likelihood of occurring among the data. Thus, each label $l \in \Lambda$ has weight $1/L$.
    \item For a fixed label $l \in \Lambda$, the weight of $X_i \in \lambda\inv(l)$ is $1/\left(LN_l\right)$ where $N_l$ is the number of point clouds with label $l$ found by $|\lambda\inv(l)|$. That is to say, assume each $X_i \in \lambda\inv(l)$ is an equally likely representation of the distribution of the points with label $l$ (regardless of the number of points in $X_i$).
    \item For each $x\in X_i$, assume equal weight and thus,
        \[w(x) := \frac{1}{LN_l |X_i|}.\]
\end{enumerate}

\begin{defn}
Let $\Omega \subset \R^D$ be a compact and convex region. Let the sets $\mathcal{X},\Lambda, \underline{\mathcal{X}}, \underline{\mathcal{X}}_l$ and the functions $\lambda, w$ be defined as before.

\begin{enumerate}
    \item The total weight of $l \in \Lambda$ in $\Omega$ is defined as
        \[w_l(\Omega) = \sum_{x \in \Omega \cap \underline{\mathcal{X}}_l} w(x).\]
    \item The total weight of $\Omega$ is defined as
        \[W(\Omega) = \sum_{l \in \Lambda} w_l(\Omega).\]
\end{enumerate}
\end{defn}

\begin{defn}
    For any compact and convex region $\Omega \subset \R^D$, define its \textbf{entropy} as
    \[S(\Omega) = - \sum_{l \in \Lambda} \frac{w_l}{W}\log_L\left(\frac{w_l}{W}\right).\]
\end{defn}

Notice that when all $w_l$ are equal for a region $\Omega$, then entropy is maximized to 1 since $w_l/W = 1/L$. When all but one $w_l$ are equal to 0, that means $\Omega$ contains only points with label $l$, and thus the entropy is 0. The goal is to find regions $\Omega$ where this entropy quantity is minimal. CDER is an algorithm that accomplishes this task.

CDER uses cover trees to aid in the identification of compact and convex regions $\Omega \subset \R^D$ of low entropy given a collection of labeled point clouds. Cover trees are hierarchical data structures used for efficient nearest neighbor search in high-dimensional spaces. We define them briefly here based on the definition given in \citep{cover_trees}.

A \textbf{cover tree} $T$, constructed for a given data set $X$, is a hierarchical tree structure where each level serves as a ``cover" for the level immediately below it. At each level, every node is linked to a corresponding point from the data set $X$ and each level is indexed by the variable $i$. While a point in $X$ may have associations with multiple nodes within the tree, a constraint is that any point should appear at most once at each level. Denoting $C_i$ as the collection of points in $X$ connected with nodes at level $i$, the cover tree adheres to the following principles for all levels $i$:

\begin{enumerate}
    \item \textit{Nesting:} $C_i \subset C_{i-1}$
    \item \textit{Covering Tree:} For any point $p$ belonging to $C_{i-1}$, there exists a point $q$ in $C_i$ such that the $d(p,q) < 2^i$ and the node at level $i$ associated with $q$ is a parent of the node at level $i - 1$ associated with $p$.
    \item \textit{Separation:} For all distinct points $p$ and $q$ within $C_i$, $d(p, q) > 2^i$.
\end{enumerate}

The tactic of CDER is to build a cover tree while using entropy to hone in on specific regions that are distinguishing for certain labels. Once a region is identified, CDER will build a distributional coordinate on that region.

When building the cover tree, the CDER algorithm can control the granularity of the covers via an optional parameter (``parsimonious") \citep{cder}.  The authors of the CDER algorithm note that, ``For machine learning applications, the “parsimonious” version [of CDER] is far faster..." (Section 6 of \citep{cder}). This is the version we use in our applications.

The distributional coordinates CDER outputs is what allows us to vectorize the point clouds from $\mathcal{X}$ into input for a supervised machine learning algorithm. A distributional coordinate is defined as follows. 

\begin{defn}
Let $X = \{x_1, \dots, x_n\} \subset \R^D$ be a point cloud. Given any function $g: X \to \R$, it is referred to as a \textbf{distributional coordinate} and define
\[\int_X g = w_1g(x_1)+\dots+w_ng(x_n)\]
where each $w_i \in \R$ is a weight associated with the point $x_i$. Any ordered set $(g_1, \dots, g_k)$ of distributional coordinates maps a point cloud $X$ into a $k$-dimensional vector $(\int_X g_1, \dots, \int_X g_k).$ 
\end{defn}

Specifically, in the case of CDER, the distributional coordinates are multidimensional Gaussians. For further reading into the details of the CDER algorithm, the reader is referred to the original manuscript \citep{cder}.

For an unlabeled point cloud $\tilde{X}$, one would expect that $\tilde{X}$ would receive the label of the distributional coordinate that had the highest evaluation. However, there can be multiple distributional coordinates with the same labels and it is not clear the most effective way to combine them. Thus, in our experiments in Section~\ref{sec:cder_application}, we let the supervised learning algorithm decide the labels based on the raw vector constructed by the evaluations under all distributional coordinates.

\section{Data Collection and Preprocessing}\label{sec:data_context}

\subsection{Data Sources and Background}

In \citep{rocklin}, the authors generate and experimentally measure the stability of synthetic mini-proteins with four distinct secondary-structure topologies. We will refer to the four distinct protein topologies as HHH, EHEE, HEEH, and EEHEE each characterized by varying levels of complexity, where H indicates an $\alpha$ helix and E represents a $\beta$ sheet.  Among these, the HHH topology exhibits relatively modest intricacy, involving just two loops and localized secondary structures, primarily helices. On the other hand, the EEHEE fold necessitates four loops and encompasses a mixed arrangement of parallel and antiparallel beta-sheets \citep{Jakubowski2023Secondary}.

As a note, we have two notions of topology being used in this manuscript that should not be confused. One notion refers to the secondary structure of the designed proteins as understood by biologists. The other notion is that of homological descriptions of the connectedness, loop-structure, and void-structure of the 3D conformation of the protein as captured in a PD. The context will make clear which notion of topology is being discussed. Usually we will use ``topology'' to mean the designed secondary sequence of the protein and we use the term ``topological'' when referencing the mathematical concept.

The approach in \citep{rocklin} involved the \textit{de novo} design of tens of thousands of proteins for each topology using a blueprint-based strategy. Each design was characterized by a unique three-dimensional backbone conformation and a sequence was predicted to be highly compatible with that specific conformation \citep{rocklin} using the Rosetta design platform \citep{alford2017rosetta}. 


To rigorously assess the stability of each protein, a stability assay using titration by both chymotrypsin and trypsin was deployed to measure an ec50 value reflecting resistance of each protein to enzymatic degradation. A ranking system based on the lower stability score obtained from either trypsin or chymotrypsin was used to assign an ec50 value to each protein. To determine a final measure of stability, an unfolded state model was built to predict a baseline resistance to enzymatic degradation by training on stabilities from randomized amino acid sequences. This stability score represents a fold change in the measured ec50 value over the model inferred baseline. The baseline unfolded state model reported in \citep{rocklin} was improved using a convolutional neural network unfolded state model defined in \citep{plos_proteins}. It is the latter, and the associated final stability scores, used in the present study.

\begin{figure}[ht]
    \centering
    \includegraphics[width=\columnwidth]{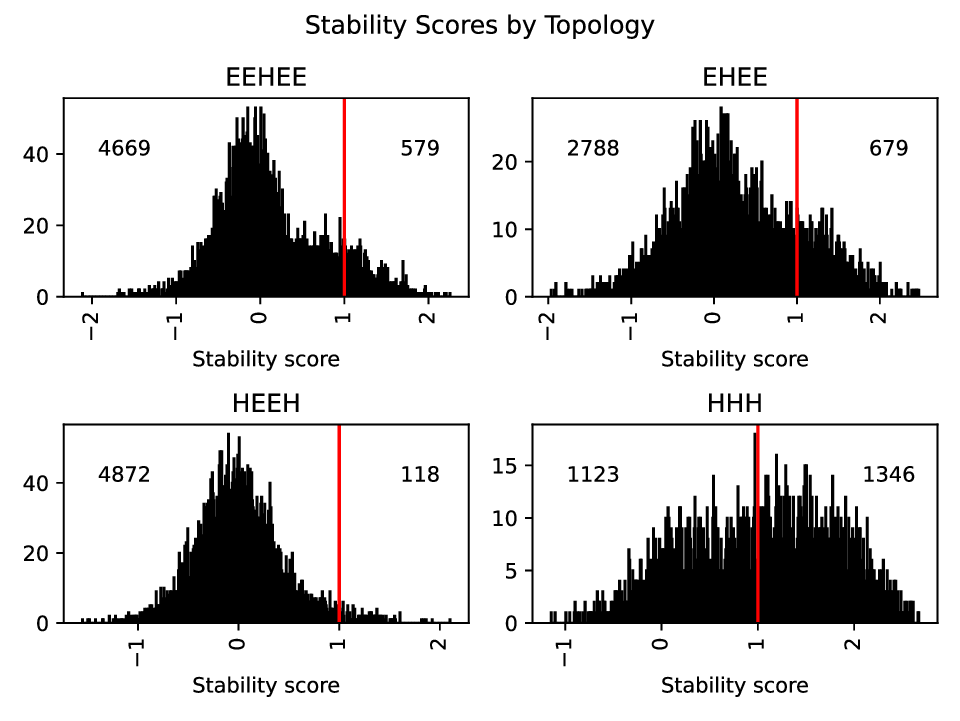}
    \caption{Histograms of distribution of stability scores by topology. The numbers in the upper left/upper right of each frame display the number of stability scores that are less/greater than 1, respectively.}
    \label{fig:stability_scores_dist}
\end{figure}

\subsection{Data Preprocessing}\label{sec:data_preprocess}

Our dataset consisted of 16,174 synthetic mini-proteins---designed and relaxed using Rosetta \citep{alford2017rosetta}---together with experimentally determined stability scores. Associated to each design is a Protein Data Bank (PDB) \citep{pdb_citation}  file containing the design's atomic coordinates and identities. The distribution of stability scores is illustrated in Figure~\ref{fig:stability_scores_dist}. There were 5248 proteins with EEHEE secondary structure topology with 579 of them with stability score above 1. Similarly, there are 3467 EHEE proteins with 679 of them with stability score above 1. There are 4990 HEEH proteins with 118 scores above 1. There are 2469 HHH proteins with 1346 scores above 1.

\begin{figure}[ht]
    \centering
    \includegraphics[width=\columnwidth]{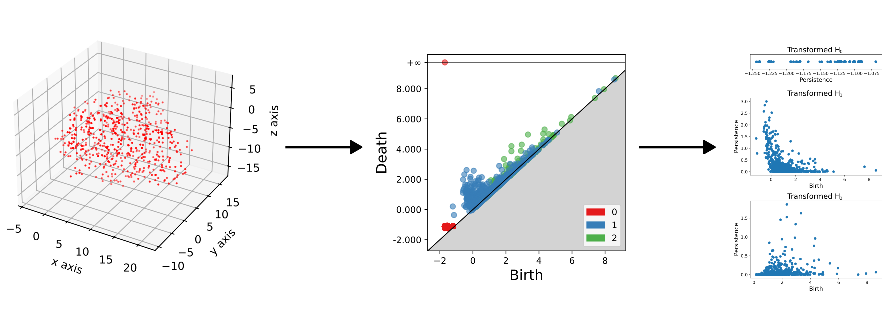}
    \caption{The atomic coordinates of a sample protein with EEHEE secondary structure topology are on the left, $H_0$, $H_1$, and $H_2$ persistence diagrams are in the middle, and the transformed diagrams are on the right. The transformed $H_0$ diagram has points distributed along one dimension because all points have the same birth. The units on all axes are angstroms (Å).}
    \label{fig:sample_protein}
\end{figure}

Using GUDHI's Weighted Alpha complex library \citep{gudhi:AlphaComplex}, we processed the PDB files into persistence diagrams using the atomic coordinates as coordinates in $\R^3$ and the van der Waals radius of each atom as the weights (H=1.2 \r{A}, N=1.55 \r{A}, O=1.52 \r{A}, C=1.7 \r{A}, S=1.8 \r{A}). Next, we transformed the birth-death persistence pairs of each diagram separately (by homology dimension) into their corresponding birth-persistence pairs using the transformation $(b,d) \mapsto (b, d-b)$ for $H_1$ and $H_2$ diagrams and $(b,d) \mapsto (d,0)$ for $H_0$ diagrams. An example of this pipeline can be seen in Figure~\ref{fig:sample_protein}.

\begin{figure}[ht]
    \centering
    \includegraphics[width=0.7\columnwidth]{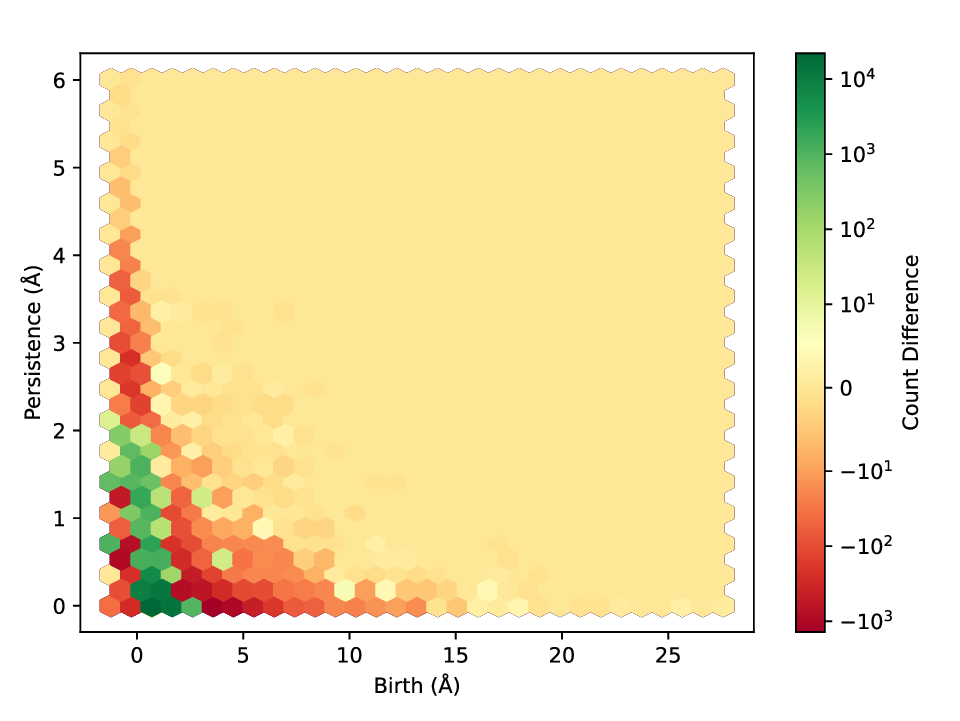}
    \caption{Hexbin plot of $H_1$ persistence pairs from the transformed persistence diagrams of all proteins with secondary structure EEHEE. Each hexagon is colored based on the number of points in that region that correspond to proteins labeled stable (green) or unstable (red). The color scale is logarithmic with deep red corresponding to a negative difference in the counts of the points corresponding to stable/unstable proteins (i.e. the number of points from persistence diagrams corresponding to unstable proteins outnumbered the number of points corresponding to stable proteins.) Similarly, deep green signifies that points corresponding to stable proteins outnumber the points corresponding to unstable proteins in that region. Yellow regions signify an equal number of points corresponding to stable and unstable proteins.}
    \label{fig:hex_plot}
\end{figure}

Previously it had been observed that the most and least stable proteins in this dataset appeared to be characterized by different numbers of $H_2$ persistence pairs, with the most stable proteins having more small scale voids and the most unstable designs having higher persistence voids \cite[Fig. 2.]{mottahyperparam2019}. To confirm these observations and more systematically visualize the distribution of persistence pairs of stable/unstable proteins, we examined the proteins with secondary structure topology EEHEE. We took all of the 579 proteins with stability score above 1 and labeled them as ``stable''; similarly, we took the 579 proteins with the lowest stability score and labeled them as ``unstable''. After computing the persistence diagrams using the pipeline described in Figure~\ref{fig:sample_protein}, we superimposed all $H_1$ diagrams on the same plot. Using a hexbin plot, we plotted the differences in points in each hexagonal region according to whether that region was dominated by points corresponding to diagrams generated from stable or unstable proteins. The plot in Figure~\ref{fig:hex_plot} shows that unstable EEHEE proteins tend to have $H_1$ persistence diagrams that dominate with points with small birth and high persistence. Physically, we expect such proteins to have larger loop structures. Figure~\ref{fig:hex_plot} also displays how there is a higher concentration of low birth and low persistence points in the $H_1$ diagrams that correspond to stable EEHEE proteins. The figure also shows yellow regions that are uninformative in discriminating between PDs corresponding to stable or unstable proteins. These regions have similar densities of persistence pairs from stable and unstable design, including many which contain no persistence pairs at all. 


\section{CDER for Feature Engineering}\label{sec:cder_application}

The stability assay and the derived stability scores are inherently noisy \citep{rocklin}, and so there is naturally ambiguity in assigning a label according a hard threshold (1) on the stability score. Moreover, for most topologies, there were significantly more proteins with a stability score below 1 than above. For these reasons, and because one of goals was to learn interpretable topological features that may correlate with stability, we first compared the topological features of the most and least stable proteins. The distributions of the stability scores for each topology is given in Figure~\ref{fig:stability_scores_dist}. Setting a stability score of 1 as the threshold between ``unstable'' and ``stable'' proteins (as done in \citep{rocklin}), we downsampled to an equal number of unstable/stable proteins up to the smaller of the two categories. That is to say, for each topology we downsampled to
\begin{enumerate}
    \item HHH: 1123 stable and 1123 unstable proteins
    \item HEEH: 118 stable and 118 unstable proteins
    \item EHEE: 679 stable and 679 unstable proteins
    \item EEHEE: 579 stable and 579 unstable proteins
\end{enumerate}

\subsection{Feature Engineering} \label{subsec:cder_feat_eng}


We used CDER to identify regions of low entropy between the diagrams associated with stable proteins and those associated with unstable proteins, separately for each secondary structure topology and each homological dimension. For each diagram type, CDER outputted a small number of labeled Gaussians. As an example, in Figure~\ref{fig:HEEH_corr}, after learning the distributional coordinates for the preprocessed $H_2$ diagrams, CDER identified two regions of low entropy marked by two labeled Gaussians (one green and one red). Roughly, that means that the green Gaussian is a region of persistence pairs that characterizes stable protein $H_2$ diagrams and vice versa for the red region. After learning these coordinates, we go through all $H_2$ diagrams of all training proteins we use for our model and evaluate the preprocessed $H_2$ diagram as a point cloud which is integrated separately against each of those two Gaussians to produce a 2-dimensional feature vector as described in Section~\ref{sec:cder_theory}. 

To be clear, care was taken to learn distributional coordinates only on diagrams chosen to be in a training set. These learned CDER distributions were then used to generate CDER features for all diagrams including those held out for model validation, 

\subsection{Model Training}

To build baseline models to compare performance of learned topological features and topological-based models, we obtained a large collection (109) of subject-matter expert (SME) features for each protein, given in Table~\ref{tab:sme_features}. Each protein was vectorized into a 109-length SME feature vector used to train stability prediction models.

To fairly assess performance of the models trained on topological features, each secondary structure topology was partitioned into a 80/20 train/validation split. 
After transforming each homological dimension persistence diagram for each protein using the  distributional coordinates (see Section~\ref{subsec:cder_feat_eng}) learned from the training data, we concatenated all CDER feature vectors together for each protein and trained a secondary-structure-topology-specific random forest classifier \citep{breiman_2001} with the stability labels of the proteins as the target variable. We used a randomized search over a specified hyperparameter space using 10-fold cross-validation to boost performance of our classifiers. 
After finding the best hyperparameters, each topology's best classifier was tested on the held out 20\% of proteins for each topology. This procedure was repeated 10 times with randomization in the train/validation split step. The results are summarized in and analyzed from Table~\ref{tab:cder_sme_rfc}. The table shows the mean APS score across 10 runs of our procedure for each featurization technique. With about $1/10^{th}$ the number of SME features, the CDER features-trained models achieve 92\%-99\% of the performance of the SME-based models.

To further see if combining SME and CDER features improved the predictive capacity of models trained on either CDER features or SME features alone, we ran the ML pipeline again, training on the combined feature sets, and report the results in the CDER+SME column of Table~\ref{tab:cder_sme_rfc} broken down by topology. A one-tailed paired t-test shows that the CDER+SME features-based models performed statistically better for the HEEH topology proteins by a margin of about 3\% improvement over the SME-based models. 

However, it is notable that combined SME and CDER models did not show significant improvements across all protein secondary structure topologies. One hypothesis for why some models trained only on CDER features perform nearly as well as the models trained on SME features, and that the combined models show little improvement over baseline, is that there is correlation between important SME and learned CDER features.

To test this theory and investigate correlations between SME features and CDER features, we omitted the train/validation split of our downsampled proteins. Since we don't need to test any models for performance in this portion of the experiment, this allowed for as much training data as possible. After transforming each homological dimension persistence diagram for each protein using the learned distributional coordinates on all of the data, we concatenated the vectors together for each protein and trained a secondary-structure-topology-specific random forest classifier with the stability labels of the proteins as the response variable. Further, we used a randomized search over a specified hyperparameter space using 10-fold cross-validation to boost performance of our classifiers. The metric for cross-validation was again average precision score (APS). After finding the best random forest classifier hyperparameters, 
we found the mean decrease in impurity within each tree of the classifier using the feature importance attribute of the random forest classifier in scikit-learn \citep{scikit-learn}. We generated feature importance of each CDER feature for training the best classifier. The correlations between the most important CDER features and the SME features were later analyzed from Figures~\ref{fig:HHH_corr} to~\ref{fig:EEHEE_corr}.

\subsection{Analysis}

\begin{table}
    \centering
    \caption{Means and standard deviations of average precision scores (APS) of 10 models trained on the respective feature set of the column and for the respective secondary structure topology of the row. The rows labeled `num\_feat' show the maximum number of features used when training 10 models for the respective topology type and feature set. The last column is the p-value of a one-tailed paired t-test assessing if the mean score of the CDER+SME model was higher than the mean score of the SME model.}
    \label{tab:cder_sme_rfc}
    \scriptsize       
    \begin{tabular}{cc||ccc|c}
        & & & & & p-val (SME vs  \\
        & & SME & CDER & CDER+SME &  CDER+SME) \\
        \hline
        HHH & APS & 0.891944 $\pm$ 0.010124 & 0.823556 $\pm$ 0.019266 & 0.891604 $\pm$ 0.009865 &  0.3428\\
        & num\_feat & 109 & 9 & 118 & \\
        \hline
        HEEH & APS & 0.836395 $\pm$ 0.062499 & 0.830680 $\pm$ 0.089472 & 0.864092 $\pm$ 0.075277 & 0.0015\\
        & num\_feat & 109 & 11 & 120 &  \\
        \hline
        EHEE & APS & 0.968058 $\pm$ 0.009562 & 0.901349 $\pm$ 0.023872 & 0.968383 $\pm$ 0.010017 & 0.2722\\
        & num\_feat & 109 & 10 & 119 &  \\
        \hline
        EEHEE & APS & 0.996776 $\pm$ 0.001877 & 0.986866 $\pm$ 0.003861 & 0.997050 $\pm$ 0.001677 & 0.0516\\
        & num\_feat & 109 & 12 & 121&  \\
    \end{tabular} 
\end{table}

Table~\ref{tab:cder_sme_rfc} shows that models trained on just the 109 SME features outperform the models trained on just the CDER features for each topology, on average. However, it is noteworthy that the models trained on just CDER features performed about 92\%-99\% as well as the models trained on just the SME features, while using about an order of magnitude fewer features. Further, these features were extracted with only knowledge of the atomic coordinates and elemental identity without the 109 biophysical/biochemical measurements determined by subject-matter experts.

When combining the CDER and SME features, the trained models performed as well or better than the models trained on SME or CDER features alone. To better understand the effect of using both CDER and SME features, we used a one-tailed paired t-test to assess the claim that the mean score of the models trained on CDER+SME features is higher than the mean score of the models trained on just the SME features. We used a paired t-test because each of the 10 models was tested on the same test set of proteins from the respective topologies.

The p-values of this test are reported in the last column of Table~\ref{tab:cder_sme_rfc}. The lowest p-value was 0.0015 for the HEEH topology proteins. This provides evidence that training our ML models on CDER+SME features perform better with respect to the mean APS than simply using the SME features for the HEEH topology proteins. Specifically, we saw about a $3\%$ increase in mean APS. Depending on one's significance level, a similar conclusion can be drawn for the EEHEE topology proteins that resulted in a p-value of 0.0516. However, it is important to also acknowledge that our experiments do not provide conclusive evidence if combining CDER and SME features results in improved performance for other topologies. Further, we do not know what the performance would be if we used all proteins instead of downsampling to the most and least stable proteins.

The code for this experiment is available for reproducing results and figures \citep{cder_classification}. The data is available upon request.

\begin{figure}[ht]
    \centering
    \includegraphics[width=\columnwidth]{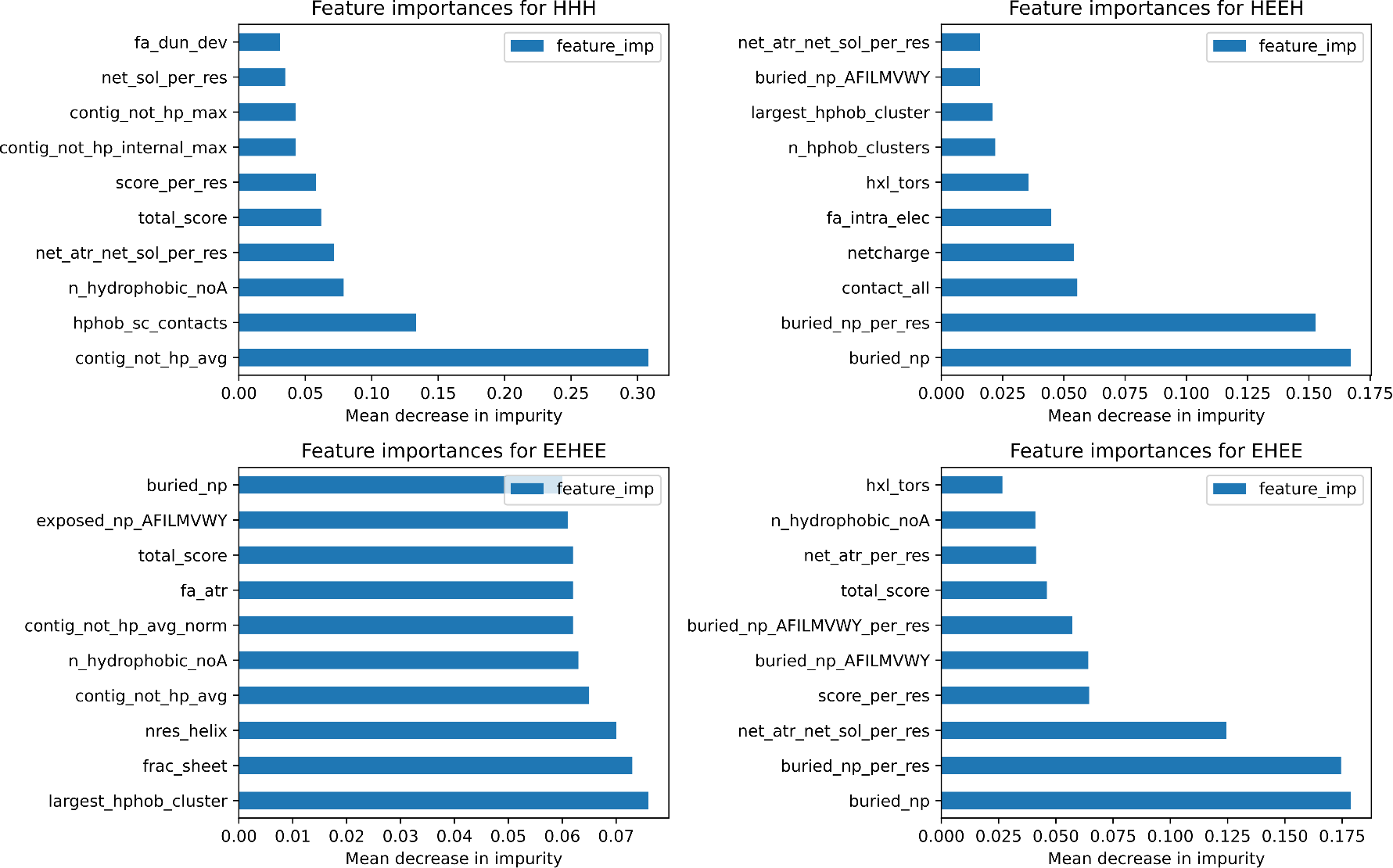}
    \caption{Top 10 feature importances in ascending order of mean decrease in impurity when training a random forest classifier on all downsampled proteins for each topology. There was no train/test split done in generating this figure. (For full descriptions of the SME features, see Table~\ref{tab:sme_features} in Section~\ref{sec:sme_feat_descrip}.)}
    \label{fig:feature_imp}
\end{figure}

\begin{figure}[]
    \centering
    \includegraphics[width=\columnwidth]{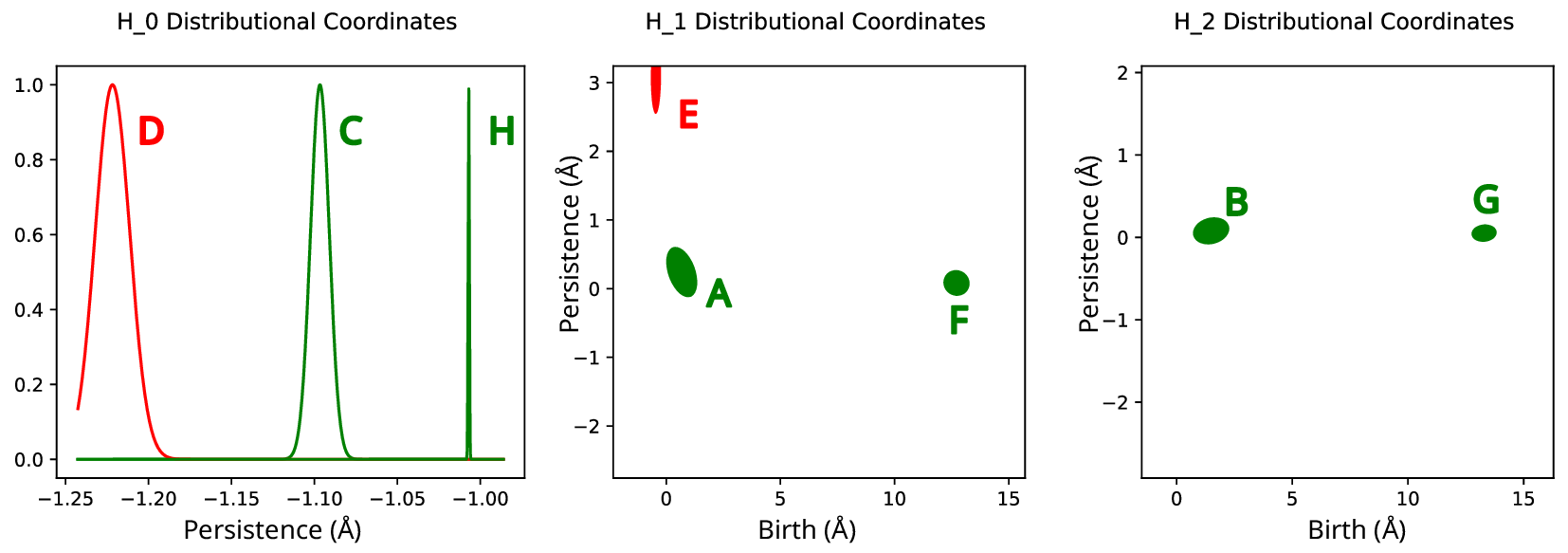}
    \hrule
    \includegraphics[width=\columnwidth]{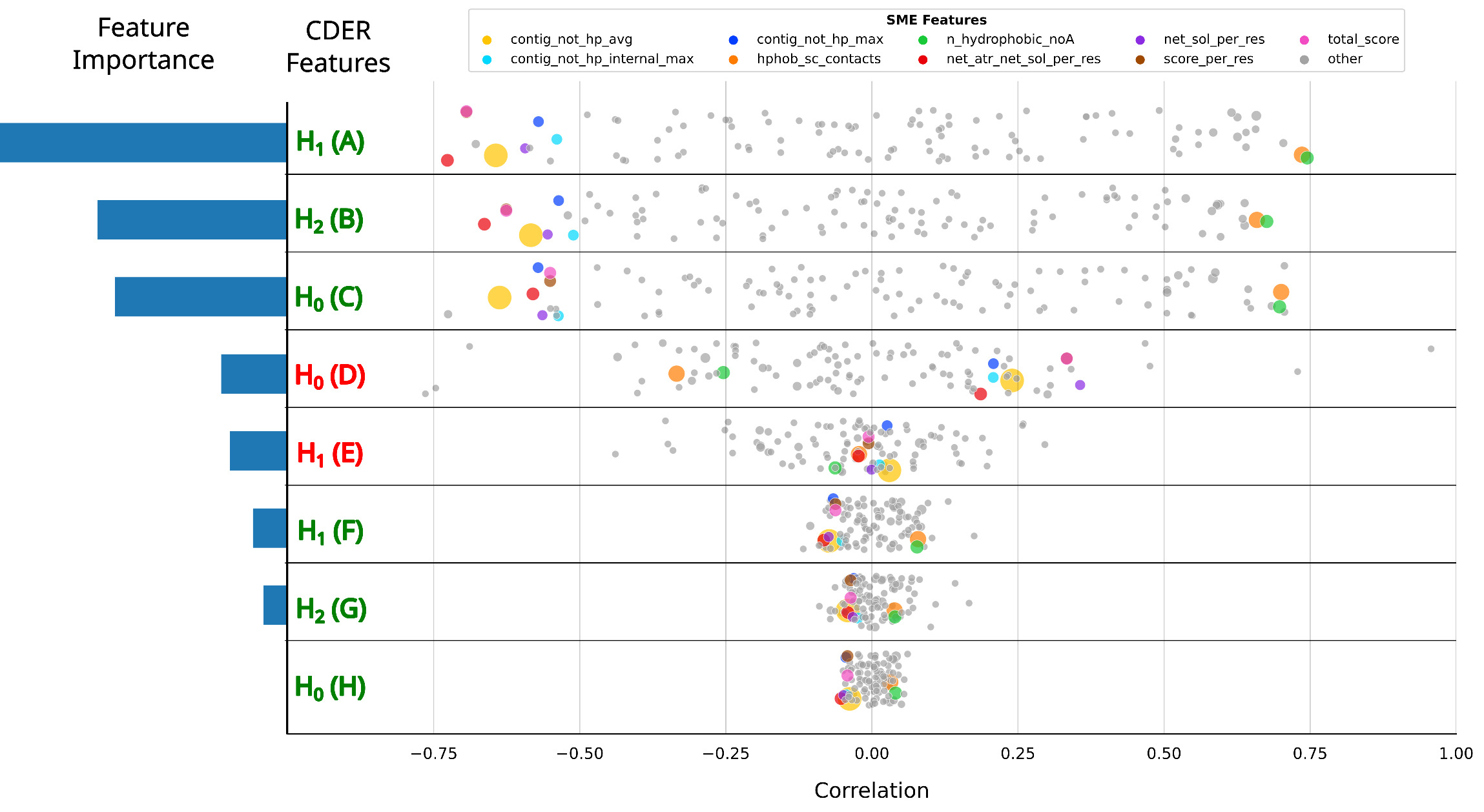}
    \caption{(Top) CDER distributional coordinates learned from persistence diagrams of proteins with HHH secondary structure topology. (Bottom) Scatter plot showing Pearson correlations between SME features and CDER features after training random forest classifier on all 2246 HHH topology proteins. The feature importance of CDER features in the CDER features-trained model is represented by the bar graph on the left side. The CDER features are labeled by color and letters to correspond with the distributional coordinates in the figures on the top. The feature importance of SME features in the SME features-trained model is represented by the size of the circle markers. Only the top 9 most important SME features are labeled in the legend.  (For full descriptions of the SME features, see Table~\ref{tab:sme_features} in Section~\ref{sec:sme_feat_descrip}.)}
    \label{fig:HHH_corr}
\end{figure}

\begin{figure}[]
    \centering
    \includegraphics[width=\columnwidth]{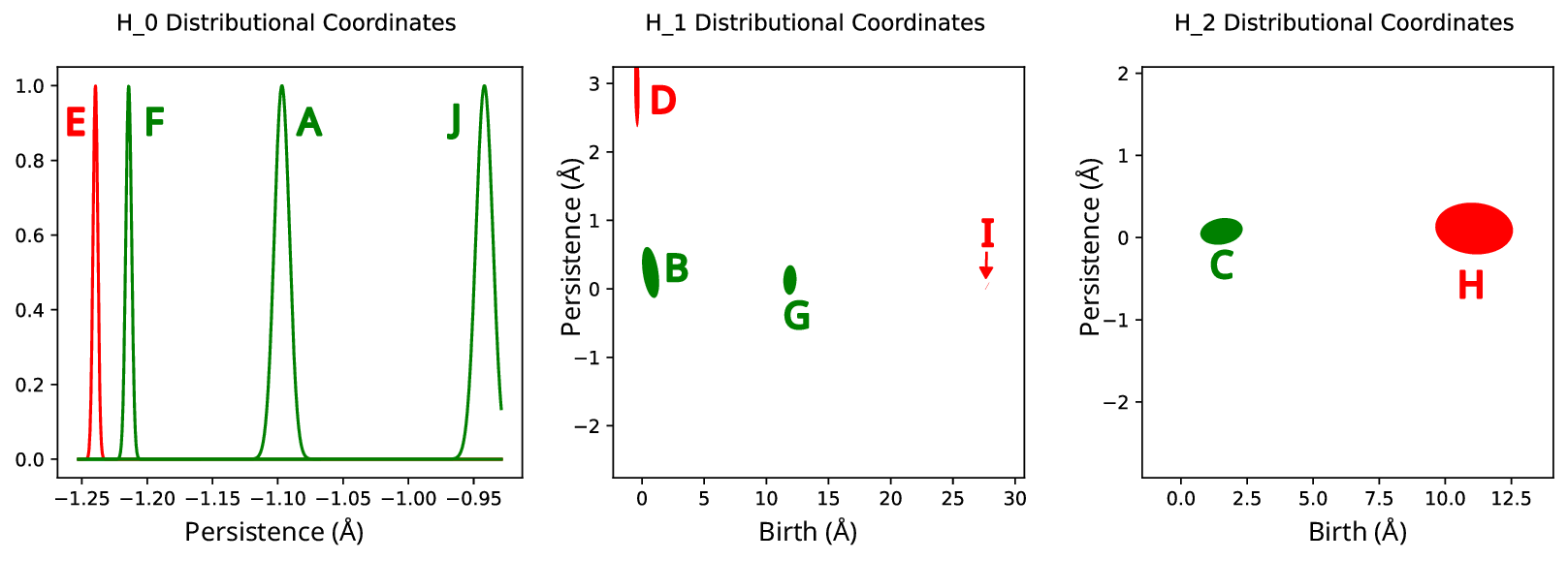}
    \hrule
    \includegraphics[width=\columnwidth]{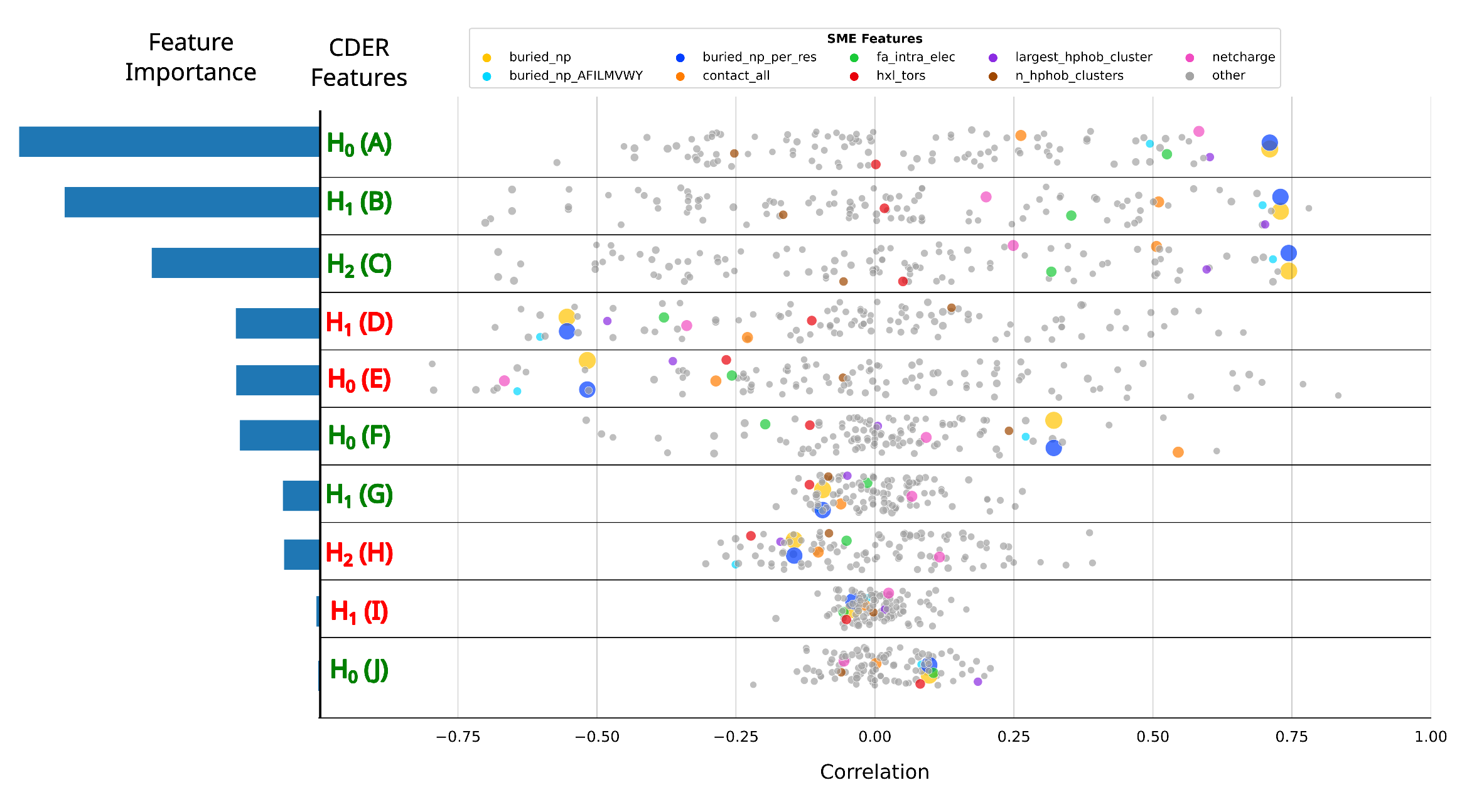}
    \caption{(Top) CDER distributional coordinates learned from persistence diagrams of proteins with HEEH secondary structure topology. (Bottom) Scatter plot showing Pearson correlations between SME features and CDER features after training random forest classifier on all 236 HEEH topology proteins. The feature importance of CDER features in the CDER features-trained model is represented by the bar graph on the left side. The CDER features are labeled by color and letters to correspond with the distributional coordinates in the figures on the top. The feature importance of SME features in the SME features-trained model is represented by the size of the circle markers. Only the top 9 most important SME features are labeled in the legend.  (For full descriptions of the SME features, see Table~\ref{tab:sme_features} in Section~\ref{sec:sme_feat_descrip}.)}
    \label{fig:HEEH_corr}
\end{figure}

\begin{figure}[]
    \centering
    \includegraphics[width=\columnwidth]{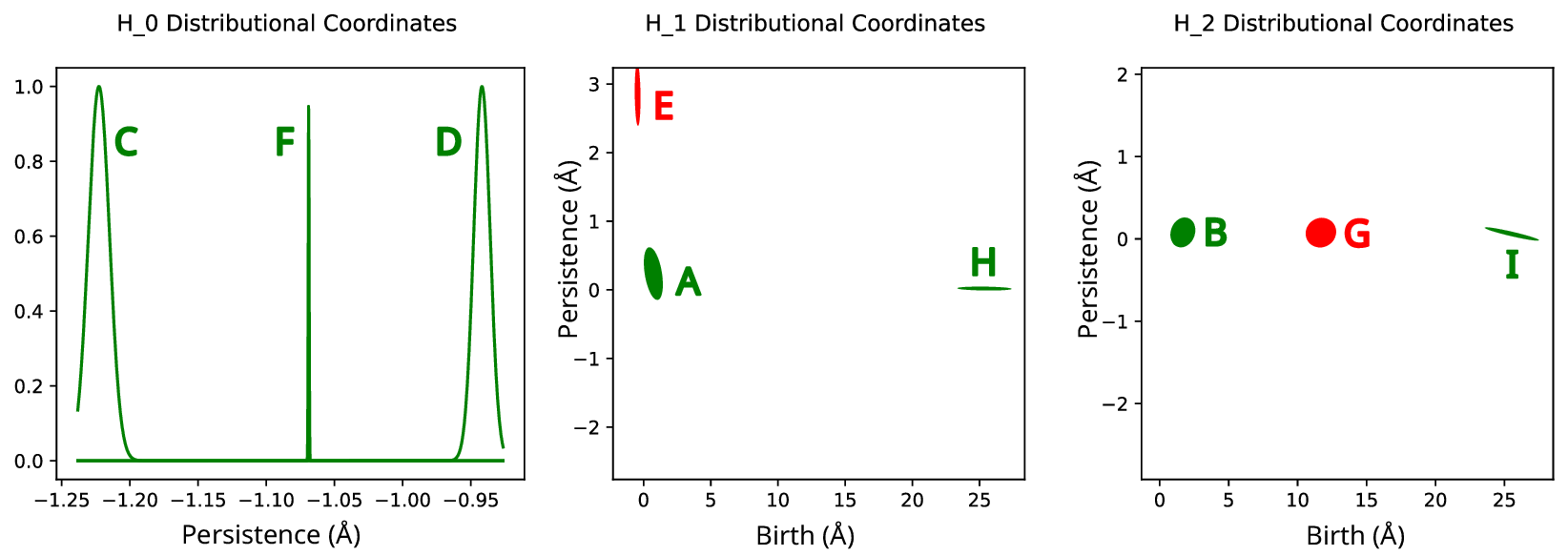}
    \hrule
    \includegraphics[width=\columnwidth]{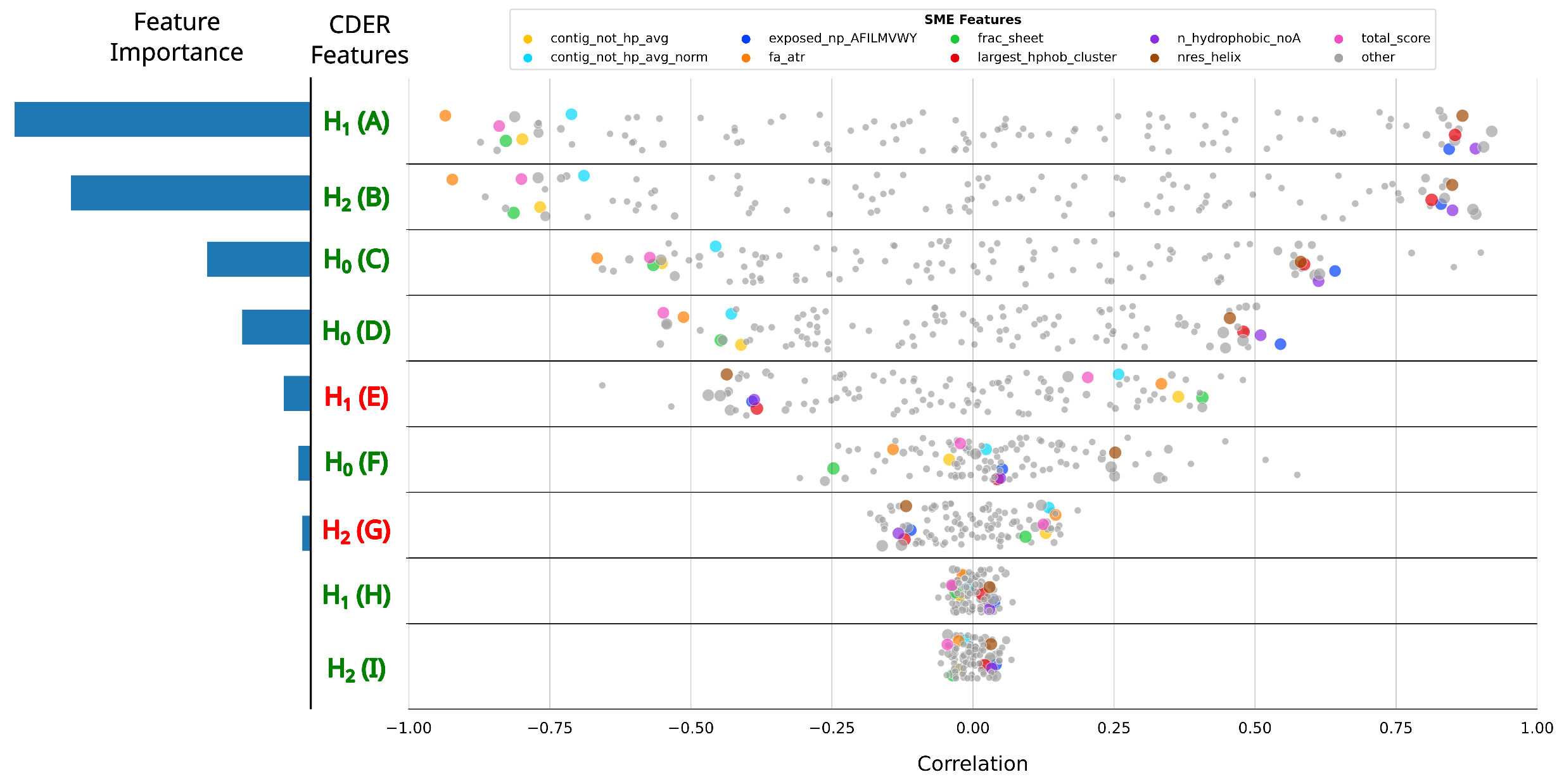}
    \caption{(Top) CDER distributional coordinates learned from persistence diagrams of proteins with EEHEE secondary structure topology. (Bottom) Scatter plot showing Pearson correlations between SME features and CDER features after training random forest classifier on all 1158 EEHEE topology proteins. The feature importance of CDER features in the CDER features-trained model is represented by the bar graph on the left side. The CDER features are labeled by color and letters to correspond with the distributional coordinates in the figures on the top. The feature importance of SME features in the SME features-trained model is represented by the size of the circle markers. Only the top 9 most important SME features are labeled in the legend.  (For full descriptions of the SME features, see Table~\ref{tab:sme_features} in Section~\ref{sec:sme_feat_descrip}.)}
    \label{fig:EEHEE_corr}
\end{figure}

\begin{figure}[]
    \centering
    \includegraphics[width=\columnwidth]{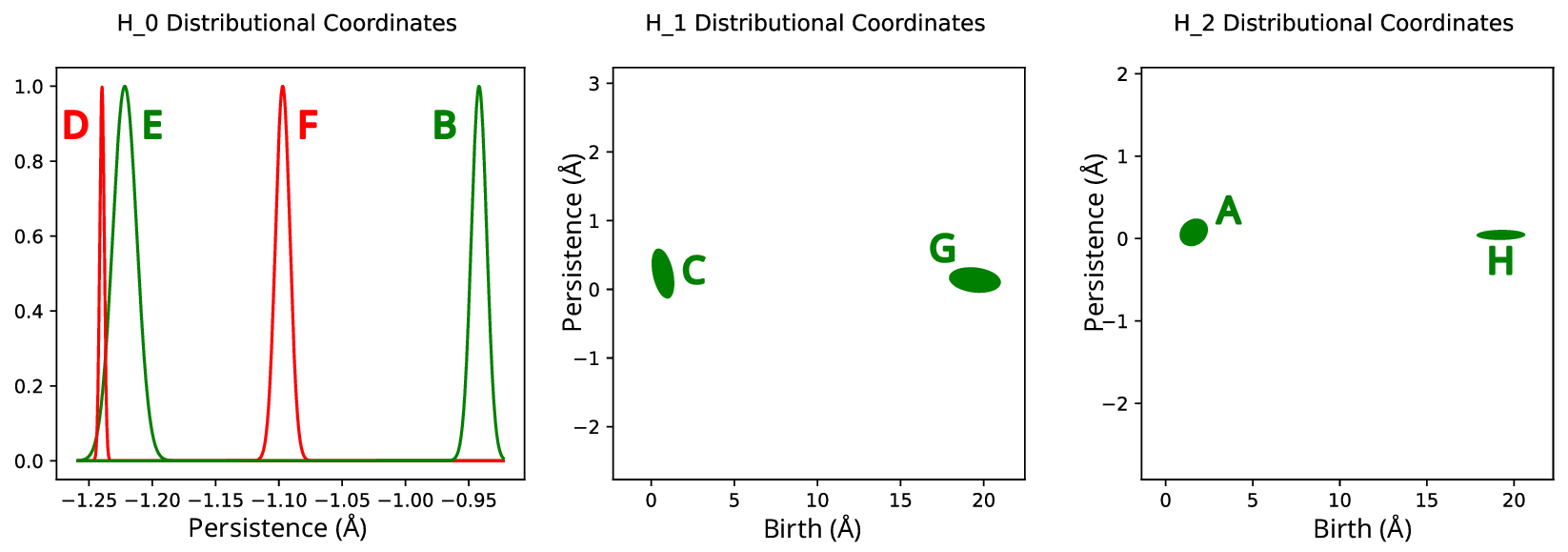}
    \hrule
    \includegraphics[width=\columnwidth]{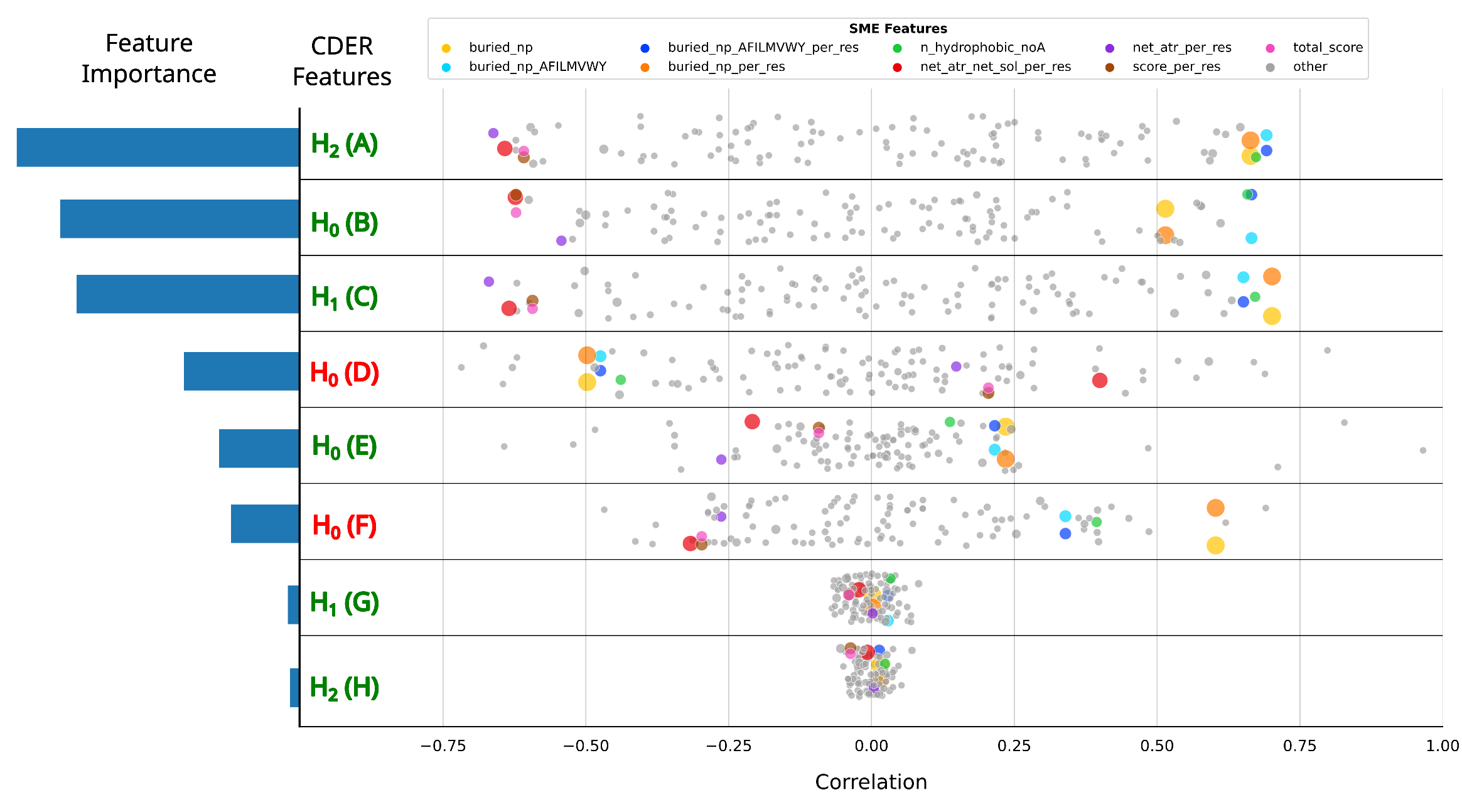}
    \caption{(Top) CDER distributional coordinates learned from persistence diagrams of proteins with EHEE secondary structure topology. (Bottom) Scatter plot showing Pearson correlations between SME features and CDER features after training random forest classifier on all 1358 EHEE topology proteins. The feature importance of CDER features in the CDER features-trained model is represented by the bar graph on the left side. The CDER features are labeled by color and letters to correspond with the distributional coordinates in the figures on the top. The feature importance of SME features in the SME features-trained model is represented by the size of the circle markers. Only the top 9 most important SME features are labeled in the legend.  (For full descriptions of the SME features, see Table~\ref{tab:sme_features} in Section~\ref{sec:sme_feat_descrip}.)}
    \label{fig:EHEE_corr}
\end{figure}

The plots in Figures~\ref{fig:HHH_corr},~\ref{fig:HEEH_corr},~\ref{fig:EHEE_corr},~\ref{fig:EEHEE_corr} each contain a plethora of information related to a single secondary structure topology and its associated models in our experiments. In particular each figure shows Pearson correlations between SME and CDER features, SME feature importance determined by SME-only models, and CDER feature importance determined in CDER-only models. Although the size of the circle markers on the correlation scatter plots indicate the importance of the SME features in the model trained on only SME features, for a more accurate and readable summary of SME feature importance we have included Figure~\ref{fig:feature_imp}. 

Looking across all plots in Figures~\ref{fig:HHH_corr}-\ref{fig:EEHEE_corr}, we can make some general observations of the data. First, we note that the CDER features of highest importance (in the model trained using only CDER features) are also highly correlated with the SME features of highest importance (in the models trained using only the SME features). In fact, there appears to be additional correlation between CDER feature importance and the maximum correlations between CDER and SME correlations. This is visible in the ``cone'' shape of the scatter plots, with the best example being Figure~\ref{fig:EEHEE_corr}. This may explain why the models based on only CDER features perform comparatively well to the models based on only SME features, as see in Table~\ref{tab:cder_sme_rfc}.

Second, we notice that across all four figures, the three CDER features with highest importance always include a feature based on a distributional coordinate from each homological dimension. Although these features are deemed important according to the random forest models, they may not all be simultaneously critical to the models' training. Explorations have been done into the effects of correlated predictors on feature importance and have identified significant  changes in feature importance when accounting for correlated predictors \citep{gregorutti2017correlation,chavent:hal-03483385}. Upon closer inspection of our own data, we find that there is correlation between the top three CDER features for each topology that ranges from 0.493 to 0.947. In fact, in Figure~\ref{fig:EEHEE_corr}, the top two CDER features labeled A and B have a 0.947 correlation. Although an accurate assessment of feature importance may be obscured by correlations between features, it is at least evident that our models find distinguishing features between stable and unstable proteins based on their connectedness ($H_0$ diagrams), loop-structure ($H_1$ diagrams), and/or void-structure ($H_2$ diagrams).

Observing Figure~\ref{fig:HHH_corr} for the HHH secondary structure topology, we see that the most important SME feature (\texttt{contig\_not\_hp\_avg}) is highly negatively correlated ($-0.644 < r < -0.584$) with the top three most important CDER features, which also correspond to regions of the persistence diagrams that are identified with stable proteins. The same three CDER features are also highly positively correlated ($0.676 < r < 0.745$) with \texttt{n\_hydrophobic\_noA}. This suggests that if a protein has a small average size of the contiguous regions of the designed sequence lacking a large hydrophobic residue, then the persistence diagrams associated with that protein shows larger values of the features generated using the distributional coordinates A, B, and C (graphed in the top of Figure~\ref{fig:HHH_corr}). Region A is associated with loop structures in the atomic arrangements that are born at relatively small length scales  and die at relatively small length scales (0-1{\AA}). Region B refers to small scale (0-1{\AA}) void structures. Thus, a protein with a lower score of \texttt{contig\_not\_hp\_avg} will also have relatively high values in the CDER features based on the regions of the persistence diagram that are associated with stable proteins (e.g., regions A, B) that are themselves indicative of small-scale loop and hole structures. Similarly, the larger the number of hydrophobic residues (F, I, L, M, V, W, and Y) in the design, the larger the values of the topological features that correspond to stable proteins, which may indicate a large presence of smalls scale loop and hole structures in the atomic arrangements of the relaxed tertiary structure.

For the HEEH secondary structure topology, the top three most important CDER features (Regions A, B, and C, also corresponding to stable proteins) are highly correlated ($0.710 < r < 0.744$) with the SME feature \texttt{buried\_np}, Figure~\ref{fig:HEEH_corr}. The latter two features again capture regions of the persistence diagrams which encode small-scale loops and voids in the protein tertiary structure, as was observed for HHH designs Figure~\ref{fig:HHH_corr}). There is also a large \textit{negative} correlation ($-0.554 < r < -0.517$) between \texttt{buried\_np} and the CDER features corresponding to unstable proteins (labeled D and E in Figure~\ref{fig:HEEH_corr}). Feature D encodes for a region of large persistence (3\AA) corresponding to larger-scale loops than those contributing significantly to feature B, for example. Thus, the larger the buried nonpolar surface area of a designed protein, apparently the higher/lower the values of the topological features that correspond to stable/unstable protein. This suggests a smaller buried nonpolar surface area will be associated with larger scale loops in the relaxed design, while larger buried nonpolar surface area correlates with a greater number of small-scale topological holes (B) and voids (C).

This is consistent with the results in \citep{rocklin}. In the section ``Global determinants of stability'', the authors note that, ``The dominant difference between stable and unstable $\alpha\alpha\alpha$ designs was the total amount of buried nonpolar surface area (NPSA) from hydrophobic amino acids. Stable designs buried more NPSA than did unstable designs.'' It is important to note that our results are for $\alpha \beta \beta \alpha$ topologies, so the comparison is not direct.

It is notable that across secondary structure topologies there are some qualitative consistencies in learned topological structures. In particular, CDER identified high-persistence regions of the $H_1$ persistence diagram as associated with unstable proteins (feature E in Figure~\ref{fig:EEHEE_corr}, feature D in Figure~\ref{fig:HEEH_corr}, and feature E in Figure~\ref{fig:HHH_corr}). This suggests a propensity for large scale loops to appear in the least stable proteins. Additionally, across all secondary structure topologies, among the top 3 most important topological determinants of stability in the topology-only models were regions of the $H_1$ and $H_2$ persistence diagrams corresponding to small-scale loops and voids, which CDER identified as associated with stable proteins. This suggests a tendency for the most stable proteins to exhibit a relatively larger number of small scale topological structures, as captured by the persistent homology of the weighted alpha filtration applied to the proteins' atomic point cloud.  

Interestingly, the most important SME features of HEEH designs appear, in general, to be less highly correlated with the most important CDER features (Figure~\ref{fig:HEEH_corr}) than for other secondary structure topologies. For instance, four or five of the top 9 most important SME features have absolute correlations near or less than 0.5 with the top three CDER features for HEEH proteins. On the other hand, all of the top 9 SME features in the other designs exhibit absolute correlations greater than 0.5 with the top three CDER features (Figures~\ref{fig:HHH_corr}, ~\ref{fig:EHEE_corr}, and ~\ref{fig:EEHEE_corr}).   This may explain why adding the CDER features to the SME feature set improved model performance most significantly for HEEH topology proteins as observed in Table~\ref{tab:cder_sme_rfc}, and less so for other topologies. However, stability prediction models of EEHEE designs also appeared to benefit marginally by combining CDER and SME features (Table~\ref{tab:cder_sme_rfc}), despite there being very high correlations between the most important SME and CDER features (Figure~\ref{fig:EEHEE_corr}).

Indeed, the EEHEE secondary structure topology, Figure~\ref{fig:EEHEE_corr} shows very high correlations ($0.689 < |r| < 0.936$) between the top two most important CDER features and all 9 labeled SME features of high importance. Particularly, we see negative correlation ($-0.936 < r < -0.666$) between \texttt{fa\_atr}, which measures the total attraction energy between atoms in the protein (Table~\ref{tab:sme_features}), and the top three most important CDER features. Since attraction energy is measured with negative values (more negative indicating greater attraction), the lower \texttt{fa\_atr} is for a design (more attractive energy), the higher the value of the CDER features that correspond to stable proteins. Remarkably, in EEHEE designs, the CDER features corresponding to small scale loops (A) and voids (B) actually exhibit a negative correlation of over -0.9 with \texttt{fa\_atr}. 


For the EHEE secondary structure topology, Figure~\ref{fig:EHEE_corr} shows a high positive correlation ($0.514< r < 0.701$) between the top three most important CDER features and each SME feature \texttt{buried\_np}, \texttt{buried\_np\_AFILMVWY}, \texttt{buried\_np\_AFILMVWY\_per\_res}, \texttt{buried\_np\_per\_res}, and \texttt{n\_hydrophobic\_noA}. Thus, for a design with a higher value for one of the SME features, we expect to see that design also have higher values for the CDER features that correspond to stable proteins. There is also high negative correlation ($-0.642 < r < -0.623$) between the top three most important CDER features and \texttt{net\_atr\_net\_sol\_per\_res}.

\section{Discussion}\label{sec:discussion}

In this work we deployed a method of data-driven feature engineering which produced parsimonious sets of interpretable topological features of protein tertiary structures which were highly correlated with protein stability measures. Our results strongly suggest that topological features may be, on their own, highly informative in modeling the stability of proteins with secondary structure HHH, EHEE, HEEH, and EEHEE. In the context of pre-defined categories for stable/unstable proteins, the data-driven features achieve classifier performances between 92\%-99\% of those models trained on a larger sets of biophysical features that were hand-curated by subject-matter experts and known to be correlated with stability. We discovered that incorporating topological/structural features with biophysical features was, in some cases, able to improve prediction of protein stability.

Although persistence diagrams are the native representation of the topological structures in a data set, they are usually transformed into finite-dimensional feature vectors before being used in machine learning modelling \citep{Pun2018PersistentHomologyBasedML}. PD ``featurization'' in which diagrams are transformed into feature vectors of fixed size, is an active area of research, with many methods appearing in the literature \citep{pers_land, Reininghaus,Bendich, adams2017persistence, Chung_frontiers_hr_ml, poly_pers_diagram}. However, these methods are not data-driven, in the sense that the transformations from PD to vector are either fixed and independent of the particular distributions of features in the training PDs, or parameterized in a limited way which can lead to information sparse features \citep{adams2017persistence}. Template systems can offer more parsimonious, flexible, and information-rich vectorizations of PDs \citep{perea2023template}. We have shown how CDER can be used to automatically learn the most discriminating regions of PDs in a challenging machine learning context. This approach represents a general TDA-ML pipeline to identify discriminating, interpretable topological features. Although CDER was proposed as a supervised classification method in its own right, we have shown how the learned distributional coordinates it outputs can be effectively used to train downstream classifiers. 

Across protein secondary structure types, CDER captured similar differential densities in the distribution of scales of topological structures between stable and unstable mini-proteins. In particular, we observed that the most stable proteins exhibited greater numbers of small-scale void and/or loop structures, while unstable proteins in certain secondary structure topologies were associated with distributional coordinates that occupy regions of high persistence loops. Furthermore, we observed high correlations between CDER features that were of high importance (in the CDER-only models) with SME features that were of high importance (in the SME-only models). This provides some explanation as to why the learned topological features train models that classify the proteins nearly as well as the models trained only on SME features, and why combining feature sets does not necessarily provide large improvements over either model individually. 



There are several important limitations of this study. First, its conclusions are limited to the small number of designed topologies we investigated. It will be important to apply our approach to a more diverse set of mini proteins representing more complex secondary structures, as that data becomes available. Second, the learned topological features reported in this study were based on structures present in the most and least stable designs. Although applying CDER to all available proteins produces qualitatively similar distributional coordinates as were reported here, and does not meaningfully alter the conclusions of the study, we have not fully explored the range of topological features that may be associated with proteins exhibiting more ambiguous stability scores. 

Since CDER is natively a classifier, and requires class labels to learn distributional coordinates, its usefulness to regression tasks has not been established. The stability assay used to characterize protein stability \citep{rocklin} produces a quantified measure, and no protein is inherently completely stable or completely unstable. We relied on reasonable thresholds of stability to assign labels, reduce the problem to a binary classification task, and compared the most and least stable designs, but other approaches could conceivably be used to develop regression models using CDER as an adaptive template system to vectorize PDs. For instance, additional thresholds could be used to bin proteins into more refined stability categories, and/or features learned to discriminate between classes could be used to train downstream regression models designed to predict actual stability scores. 

One could also pursue developing better performing and pruned models for the binary classification problem. By using methods of feature elimination/selection, correlations and feature importance analyses could be made with greater clarity. 

Finally, the topological features engineered in this study capture very simple structural characteristics of a protein. This allows for easy interpretation, but may limit the capacity of topology-based models to simple proteins and/or easier prediction tasks. Neglected from this study is any investigation into which atoms in each design are responsible for generating the structural features that contribute most to the discriminating topological features used to train our classifiers. Although the generators of persistent homology features are not unique, a number of methods exist to find representatives of topological structures with desirable characteristics (e.g., smallest path length, volume, etc.) \citep{mincyclerep}. It is conceivable that investigating specific generators of homology, say in highly unstable designs, could inform the design process or lead to biological insights. These are open directions for future work.


\acks{This manuscript contains some results previously published in AM's dissertation \citep{mydissertation}. 


The authors declare no competing interests in relation to this work.}

\appendix

\section{Persistent Homology with Data}\label{sec:extra_persistence}

In the pursuit of topological features derived from point cloud data, the underlying assumption is that the points represent samples from a topological space of interest. The structure of this space aids in distinguishing between different topological spaces. 

To infer the topological space from the finite set of sampled data points, one typically considers the data points as vertices of a simplicial complex. This complex encompasses vertices, edges (pairs of vertices), triangles (sets of three vertices), tetrahedra (sets of four vertices), and their higher-dimensional counterparts (sets of \(n > 4\) vertices). A simplicial complex provides a framework with well-defined notions of "shape." With this structured representation of the point cloud, algebraic tools are employed to compute topological descriptors of interest. Specifically, persistent homology groups are calculated to characterize topological invariants such as connected components, holes, and voids across different scales. This section provides a brief overview of these concepts and examples. For a more comprehensive understanding of the mathematical foundations, readers are referred to \citep{Edelsbrunner, hatcher2002algebraic}. For a detailed computational approach with numerous practical considerations and examples, \citep{Roadmap} is recommended.

\subsection{Simplicial Complexes and Homology}

A simplicial complex $K$ on a data set with points $ P= \{v_1, v_2, \dots, v_n\}$ is a set made of subsets of $P$ that satisfies the following conditions:
\begin{enumerate}
    \item If $\sigma$ is in $K$, then all subsets of $\sigma$ are also in $K$.
    \item The intersection of any two in $K$ is either empty or another element in $K$.
\end{enumerate}

The elements of $K$ are known as simplices where a $p$-simplex is a subset of $P$ with $p+1$ points. 

For a simplicial complex $K$ over a field $\mathbb{F}$, the $p$-th chain group $C_p(K)$ is the vector space generated by the $p$-simplices in $K$. An element of $C_p(K)$ is an $p$-chain, which is a formal sum of $p$-simplices with coefficients from $\mathbb{F}$.

The boundary map $\partial_n: C_p(K) \to C_{p-1}(K)$ is defined on an $p$-simplex $[v_0, v_1, \ldots, v_p]$ as:

\[\partial_p([v_0, v_1, \ldots, v_p]) = \sum_{i=0}^{p} (-1)^i [v_0, \ldots, \hat{v_i}, \ldots, v_p],\]

where $\hat{v_i}$ indicates that the vertex $v_i$ is omitted. The $p$-th homology group $H_p(K)$ is then defined as the quotient group:
\[H_p(K) = \text{Ker}(\partial_p) / \text{Im}(\partial_{p+1}),\]
where $\text{Ker}(\partial_p)$ is the group of $p$-cycles and $\text{Im}(\partial_{p+1})$ is the group of $p$-boundaries. The boundary operator satisfies the essential property $\partial_{p} \circ \partial_{p+1} = 0$. When viewed as a group with the appropriate scalar multiplication defined over $\mathbb{F}$, $H_p(K)$ becomes a vector space with the $p$-cycles and $p$-boundaries as subspaces.

Computing $\beta_p = \dim(H_p(K))$ gives the $p$-th Betti number, which counts the number of $p$-dimensional holes in $K$ that are not in the same equivalence class. Specifically, $\beta_0$ counts the number of connected components, $\beta_1$ counts the number of loops, and $\beta_2$ counts the number of voids. The Betti numbers are of high significance due to the fact topologically equivalent spaces have the same Betti numbers (Betti numbers are topological invariants).

\subsection{Filtrations, Persistent Homology, and Persistence Diagrams}

With data, it is not clear beforehand what the proper complex $K$ should be that best approximates the underlying topological space the data may have been sampled from. Further, there may be topological invariants of interest in various complexes that can be built on top of the point cloud data. Seeking to study topological invariants across a whole family of simplicial complexes, a filtration is defined.

A filtration of a simplicial complex $K$ is a nested sequence of subcomplexes $\{K_{r_i}\}$ such that $K_0 \subseteq K_{r_1} \subseteq \cdots \subseteq K_{r_n} = K$, where $0 < r_1 <\dots<r_n$ is a sequence of scales. Persistent homology tracks the evolution of homology groups across this filtration by studying the induced sequence of linear maps between the associated homology groups

\[H_p(K_{0}) \to H_p(K_{r_1}) \to \dots \to H_p(K_{r_n}).\]

Equivalence classes (features) that persist over long scales $r_i$ are considered valuable topological information about $K$ (and by extension, about the data). By tracking the scale at which a feature is birthed, $b$, and the scale at which the feature dies/merges with another equivalence class, $d$, the information can be summarized in a persistence diagram.

A persistence diagram is a multiset of points in the extended plane, where each point $(b, d)$ represents a homological feature that is born at scale $b$ and dies at scale $d$. The persistence of a feature is $d - b$, and points further away from the diagonal $y = x$ represent more persistent features. Persistence diagrams are stable under small perturbations, making them useful for comparing shapes in a robust manner.

\subsection{Weighted Alpha Complexes}
There are numerous ways to construct a complex $K$ (and by extension, a filtration) on point cloud data \citep{Roadmap}. The method used in this paper was the weighted alpha complex. In the context of protein modeling, every atom is depicted as a ball with its weight indicating the extent of its van der Waals interactions determined by the atom type.

Given a set of weighted points $(P, w)$, where $P$ is a set of points in $\mathbb{R}^d$ and $w: P \to \mathbb{R}_{\geq 0}$ is a weight function, the weighted alpha complex is a subcomplex of the Delaunay triangulation of $P$. It consists of simplices whose corresponding Delaunay circumspheres, when scaled by their weights, do not contain any other points of $P$ in their interiors. The weighted alpha complex adapts to the local density of the data and is useful in applications such as the one in this paper. A detailed definition of the weighted alpha complex is provided in \citep{Edelsbrunner}.

\section{Table of Feature Descriptions}\label{sec:sme_feat_descrip}

\rowcolors{2}{gray!20}{white}
\footnotesize
\begin{longtable}{l p{.62\linewidth}}
    \caption{The 109 SME features used in our models and their descriptions.} \label{tab:sme_features}\\
    \rowcolor{gray!50}
    \hline
    \textbf{Feature} & \textbf{Description} \\
    \hline
    AlaCount & Number of alanine residues \\
    \hline
    T1\_absq &  Number of charged residues in the first turn of each helix, summed over all helices \\
    \hline
    T1\_netq &  Net favorable charge in the first turn of each helix, summed over all helices \\
    \hline
    Tminus1\_absq &  Number of charged residues in the last turn of each helix, summed over all
    helices \\
    \hline
    Tminus1\_netq & Net favorable charge in the last turn of each helix, summed over all helices \\
    \hline
    Tend\_absq &  \texttt{T1\_netq + Tminus1\_netq} \\
    \hline
    Tend\_netq &  \texttt{T1\_absq + Tminus1\_absq} \\
    \hline
    abego\_res\_profile & Each position $i$ in the designed structure can be classified by its ABEGO type (86), and the ABEGO types of positions $i-1$, $i$, and $i+1$ form a triad that defines the three-residue local structure at a coarse level. The \texttt{abego\_res\_profile} metric is the sum over all positions $i$ in the designed structure of $\log \left((p_{\text{aa}} | \text{abego triad})\right)/(p_{\text{aa}})$, where $(p_{\text{aa}} | \text{abego triad})$ is the frequency of the designed amino acid (from position $i$) in regions of natural proteins sharing the same ABEGO triad as the designed region centered on position $i$, and $p_{\text{aa}}$ is the overall frequency of the designed amino acid at position $i$. At each position, this score is positive when the designed amino acid is overrepresented (compared with its normal frequency) in regions of natural proteins with the same local ABEGO triad structure as the designed region, and the score is negative when the designed amino acid is underrepresented in regions of natural proteins with the same local ABEGO triad structure    \\
    \hline
    abego\_res\_profile\_penalty & Same as \texttt{abego\_res\_profile}, except summing over only positions with negative \texttt{abego\_res\_profile scores} (positions where the designed residue is typically
    underrepresented in the local structure) \\
    \hline
    avg\_all\_frags & The average RMSD of all selected fragments to their corresponding segments of the designs, in \AA. ($200 \times (n - 8)$ fragments in total for protein length $n$)    \\
    \hline
    avg\_best\_frag & The average RMSD of the lowest-RMSD fragment for each designed segment, in \AA. ($(n - 8)$ fragments in total)    \\
    \hline
    bb & \texttt{hbond\_lr\_bb / nres} \\
    \hline
    buns\_bb\_heavy & Buried polar atoms (N and O) in the protein backbone that do not make hydrogen bonds \\
    \hline
    buns\_nonheavy & Buried polar hydrogens that do not make hydrogen bonds \\
    \hline
    buns\_sc\_heavy & Buried polar atoms (N and O) in the side chains that do not make hydrogen bonds \\
    \hline
    buried\_minus\_exposed & \texttt{buried\_np - exposed\_hydrophobics} \\
    \hline
    buried\_np & Buried nonpolar surface area in the designed structure on all amino acids, computed using version1 definitions of total nonpolar surface area per residue \\
    \hline
    buried\_np\_AFILMVWY & Buried nonpolar surface area in the designed structure on nonpolar amino acids (AFILMVWY), computed using version2 definitions of total nonpolar surface area per residue \\
    \hline
    buried\_np\_AFILMVWY\_per\_res & \texttt{buried\_np\_AFILMVWY / n\_res} \\
    \hline
    buried\_np\_per\_res & \texttt{buried\_np / n\_res} \\
    \hline
    buried\_over\_exposed & \texttt{buried\_np / exposed\_hydrophobics} \\
    \hline
    chymo\_cut\_sites & Number of F, Y, W residues in the design \\
    \hline
    chymo\_with\_LM\_cut\_sites & Number of F, L, M, Y, W residues in the design \\
    \hline
    contact\_all & Number of sidechain carbon-carbon contacts in the designed structure, computed with AtomicContactCount Rosetta filter \\
    \hline
    contig\_not\_hp\_avg & Average size of the contiguous (in primary sequence) regions of the designed sequence lacking a large hydrophobic residue (FILMVWY) \\
    \hline
    contig\_not\_hp\_avg\_norm & $$\frac{{\texttt{contig\_not\_hp\_avg}}}{{{\texttt{n\_res}}}/({{1 + \texttt{n\_hydrophobic\_noA}}})}$$
    \\
    \hline
    contig\_not\_hp\_internal\_max & Size of the largest contiguous region (in primary sequence) in the designed sequence containing no large hydrophobic residues (FILMVWY), excluding the regions between the first and last large hydrophobic residues and the termini \\
    \hline
    contig\_not\_hp\_max & Size of the largest contiguous region (in primary sequence) in the designed sequence containing no large hydrophobic residues (FILMVWY) \\
    \hline
    degree & Average number of residues in a 9.5 \AA{} sphere around each residue, computed with AverageDegree Rosetta filter  \\
    \hline
    dslf\_fa13 & Disulfide geometry potential \\
    \hline
    exposed\_hydrophobics & Exposed nonpolar surface area of the designed structure, in \AA$^2$, computed using TotalSasa Rosetta filter, set to compute hydrophobic-only SASA    \\
    \hline
    exposed\_np\_AFILMVWY & Exposed nonpolar surface area in the designed structure on nonpolar amino acids (AFILMVWY) \\
    \hline
    exposed\_polars & Exposed polar surface area of the designed structure, in \AA$^2$,
    computed using TotalSasa Rosetta filter, set to compute polar-only SASA \\
    \hline
    exposed\_total & Total exposed surface area of the designed structure, in \AA$^2$,
    computed using TotalSasa Rosetta filter  \\
    \hline
    fa\_atr & Full atom attraction (Rosetta) \\
    \hline
    fa\_atr\_per\_res & \texttt{fa\_atr / n\_res} \\
    \hline
    fa\_dun\_dev & Dunbrack potential deviation energy \\
    \hline
    fa\_dun\_rot & Dunbrack potential rotational energy \\
    \hline
    fa\_dun\_semi & Dunbrack potential semi-rotational energy \\
    \hline
    fa\_elec & Electrostatic interaction energy \\
    \hline
    fa\_intra\_atr\_xover4 & Attractive interaction energy within 4 residues \\
    \hline
    fa\_intra\_elec & Intra-residue electrostatic interaction energy \\
    \hline
    fa\_intra\_rep\_xover4 & Repulsive interaction energy within 4 residues \\
    \hline
    fa\_intra\_sol\_xover4 & Solvation energy within 4 residues \\
    \hline
    fa\_rep & Repulsive interaction energy \\
    \hline
    fa\_rep\_per\_res & \texttt{fa\_rep / n\_res} \\
    \hline
    fa\_sol & Full atom solvation via Lazaridis-Karplus solvation model \\
    \hline
    frac\_helix & \texttt{nres\_helix / n\_res} \\
    \hline
    frac\_loop & \texttt{nres\_loop / n\_res} \\
    \hline
    frac\_sheet & \texttt{nres\_sheet / n\_res} \\
    \hline
    fxn\_exposed\_is\_np & \texttt{exposed\_hydrophobics / exposed\_total} \\
    \hline
    hbond\_bb\_sc & Backbone to side chain hydrogen bonds \\
    \hline
    hbond\_lr\_bb & Long-range backbone to backbone hydrogen bonds \\
    \hline
    hbond\_lr\_bb\_per\_sheet & \texttt{hbond\_lr\_bb / nres\_sheet} \\
    \hline
    hbond\_sc & Side chain to side chain hydrogen bonds \\
    \hline
    hbond\_sr\_bb & Backbone to backbone hydrogen bonds in alpha helices or turns \\
    \hline
    hbond\_sr\_bb\_per\_helix & \texttt{hbond\_sr\_bb / nres\_helix} \\
    \hline
    helix\_sc &  Average shape complementarity of each helical secondary structure element with the rest of the structure, computed using SSShapeComplementarity Rosetta filter, set to evaluate helices  \\
    \hline
    holes &  Normalized measure of the void volume inside the designed structure, computed with Holes Rosetta filter \\
    \hline
    hphob\_sc\_contacts & Total number of sidechain-sidechain contacts between large hydrophobic residues (FILMVWY) in the designed structure \\
    \hline
    hphob\_sc\_degree & \texttt{hphob\_sc\_contacts / n\_hydrophobic\_noA} \\
    \hline
    hxl\_tors & Helix-local torsion potential energy \\
    \hline
    hydrophobicity & Total hydrophobicity of the designed sequence, using the amino acid
    hydrophobicity scale from (87) \\
    \hline
    largest\_hphob\_cluster & Size of the largest group of large hydrophobic residues (FILMVWY) that are all connected by at least one contact to each other in the designed structure \\
    \hline
    lk\_ball & Anisotropic contribution to the solvation \\
    \hline
    lk\_ball\_bridge & Bonus to solvation coming from bridging waters, measured by overlap of the "balls" from two interacting polar atoms \\
    \hline
    lk\_ball\_bridge\_uncpl & Same as \texttt{lk\_ball\_bridge}, but the value is uncoupled with dGfree (i.e. constant bonus, whereas \texttt{lk\_ball\_bridge} is proportional to dGfree values) \\
    \hline
    lk\_ball\_iso & Same as \texttt{fa\_sol} \\
    \hline
    loop\_sc & Average shape complementarity of each loop element with the rest of the structure, computed using SSShapeComplementarity Rosetta filter, set to evaluate loops only \\
    \hline
    mismatch\_probability & Geometric average probability (across all positions in the design) that the designed residues will not adopt their designed secondary structures, as calculated by the PSIPRED algorithm (75) from the designed sequence. Computed using the SSPrediction Rosetta filter\\
    \hline
    nearest\_chymo\_cut\_to\_Cterm & Number of residues between the design C-terminus and the nearest F, Y, or W residue. If no F, Y, or W present, then the length of the design \\
    \hline
    nearest\_chymo\_cut\_to\_Nterm & As previously, except to the design N-terminus \\
    \hline
    nearest\_chymo\_cut\_to\_term & Minimum of \texttt{nearest\_chymo\_cut\_site\_to\_Nterm} and \text{nearest\_chymo\_cut\_site\_to\_Cterm} \\
    \hline
    nearest\_tryp\_cut\_to\_Cterm & As defined for \texttt{nearest\_chymo\_cut\_site}, except referring to K and R residues \\
    \hline
    nearest\_tryp\_cut\_to\_Nterm & As defined for \texttt{nearest\_chymo\_cut\_site}, except referring to K and R residues \\
    \hline
    nearest\_tryp\_cut\_to\_term & As defined for \texttt{nearest\_chymo\_cut\_site}, except referring to K and R residues \\
    \hline
    net\_atr\_net\_sol\_per\_res & \texttt{net\_atr\_per\_res + net\_sol\_per\_res} \\
    \hline
    net\_atr\_per\_res & \texttt{(fa\_atr + fa\_rep) / n\_res} \\
    \hline
    net\_sol\_per\_res & \texttt{(fa\_sol + fa\_elec) / n\_res} \\
    \hline
    netcharge & Net charge on the design, assuming a charge of +1 on R and K, +0.5 on H, and -1 on D and E \\
    \hline
    n\_charged & Count of D, E, K, and R residues in the designed sequence, plus one-half the number of H residues \\
    \hline
    n\_hphob\_clusters & Number of disconnected groups of large hydrophobic residues (FILMVWY), where a group is defined as residues that contact each other in the designed structure but do not contact residues outside of the group \\
    \hline
    n\_hydrophobic & Count of A, F, I, L, M, V, W, and Y residues in the design \\
    \hline
    n\_hydrophobic\_noA & Count of F, I, L, M, V, W, and Y residues in the design \\
    \hline
    n\_polar\_core & Number of polar core residues \\
    \hline
    n\_res & Number of residues in the design \\
    \hline
    nres\_helix & Number of helical residues in the design, according to DSSP \\
    \hline
    nres\_loop & Number of loop residues in the design, according to DSSP \\
    \hline
    nres\_sheet & Number of beta strand residues in the design, according to DSSP \\
    \hline
    omega & Energy from all torsional angles \\
    \hline
    one\_core\_each & Fraction of secondary structure elements (helices and strands) with one large hydrophobic residue (FILMVYW) at a position in the core layer of the designed structure \\
    \hline
    p\_aa\_pp & Compatibility between backbone torsional angles and side chain identity \\
    \hline
    pack & Normalized measure of packing density, computed using PackStat Rosetta filter \\
    \hline
    percent\_core\_SASA & \texttt{res\_count\_core\_SASA / n\_res} \\
    \hline
    percent\_core\_SCN & \texttt{res\_count\_core\_SCN / n\_res} \\
    \hline
    pro\_close & Strain energy in protein side chain conformations \\
    \hline
    rama\_prepro & Ramachandran energy for residues before Proline    \\
    \hline
    ref & Reference energy \\
    \hline
    res\_count\_core\_SASA & Number of residues in the core layer of the designed structure, with layers defined using solvent accessible surface area-based criteria \\
    \hline
    res\_count\_core\_SCN & Number of residues in the core layer of the designed structure, with layers defined using sidechain neighbors-based criteria \\
    \hline
    score\_per\_res & \texttt{total\_score / n\_res} \\
    \hline
    ss\_contributes\_core & Number of residues contributing to reference energy \\
    \hline
    ss\_sc & Average shape complementarity of each helical or loop element with the rest of the structure, computed using SSShapeComplementarity Rosetta filter \\
    \hline
    sum\_best\_frags & Sum of the RMSDs of the lowest-RMSD fragment for each designed segment. (n - 8 fragments in total) \\
    \hline
    total\_score & Overall Rosetta energy for the complete structure \\
    \hline
    tryp\_cut\_sites & Number of K, R residues in the design \\
    \hline
    two\_core\_each & Fraction of secondary structure elements (helices and strands) with two large \\
    \hline
    worstfrag & Among the set of fragments that are the lowest-RMSD fragments for their positions, the highest RMSD found \\
    \hline
    worst6frags & Among the set of fragments that are the lowest-RMSD fragments for their positions, the sum of the RMSDs of the six highest RMSD fragments \\
    \hline
\end{longtable}  

\vskip 0.2in
\bibliography{bibliography}

\end{document}